\documentclass[11pt]{article}

\usepackage[preprint]{acl}

\usepackage{times}
\usepackage{latexsym}

\usepackage[T1]{fontenc}

\usepackage[utf8]{inputenc}

\usepackage{microtype}

\usepackage{inconsolata}

\usepackage{amsmath,amsfonts,bm}

\def\eqref#1{equation~\ref{#1}}

\def\1{\bm{1}}

\DeclareMathAlphabet{\mathsfit}{\encodingdefault}{\sfdefault}{m}{sl}
\SetMathAlphabet{\mathsfit}{bold}{\encodingdefault}{\sfdefault}{bx}{n}

\DeclareMathOperator*{\argmax}{arg\,max}
\DeclareMathOperator*{\argmin}{arg\,min}

\usepackage[utf8]{inputenc} 
\usepackage[T1]{fontenc}    
\usepackage{tgtermes}
\usepackage{wrapfig}
\usepackage{tcolorbox}
\usepackage{booktabs}       
\usepackage{amsfonts}       
\usepackage{nicefrac}       
\usepackage{subcaption}
\usepackage{microtype}      
\usepackage{tabularray}
\usepackage{enumerate}
\usepackage{enumitem}
\setlist{leftmargin=1ex}
\usepackage{mathtools}
\usepackage{amsthm, amssymb}

\newtheorem{lemma}{Lemma}

\newtheorem{definition}{Definition}
\usepackage{algorithm} 
\usepackage[noend]{algpseudocode}
\usepackage{xspace}
\usepackage{multirow}
\usepackage{lipsum}
\usepackage{xcolor,colortbl}

\usepackage[normalem]{ulem}
\usepackage{mdframed}
\usepackage{xcolor}
\definecolor{babyblueeyes}{rgb}{0.63, 0.79, 0.95}
\definecolor{brightpink}{HTML}{D8315B}
\definecolor{lightpink}{HTML}{EF798A}
\definecolor{cvprblue}{rgb}{0.21,0.49,0.74}
\definecolor{myblue}{HTML}{d8ebf8}
\definecolor{lightred}{HTML}{D33E43}
\definecolor{mygreen}{HTML}{2DD881}
\definecolor{darkgreen}{HTML}{006400}
\definecolor{salmon}{HTML}{FA8072}
\definecolor{mybluee}{RGB}{0, 102, 204}

\usepackage{graphicx}
\usepackage{url}

\usepackage{caption}
\usepackage{etoolbox}
\newtoggle{showlabels}
\toggletrue{showlabels} 

\def\*#1{\mathbf{#1}}
\usepackage{graphicx}

\title{\name: Kernelized and Information Theoretic Exemplars\\ for In-Context Learning}

\def\name{\textsc{KITE}\xspace}

\newcommand{\bb}[1]{\mathbb{#1}}
\newcommand{\fl}[1]{\mathbf{#1}}
\newcommand{\ca}[1]{\mathcal{#1}}

\newtheorem{theorem}{Theorem}

\author{Vaibhav Singh$^{1}$ ~Soumya Suvra Ghosal$^{2}$ ~Kapu Nirmal Joshua$^{3}$ ~Soumyabrata Pal$^{4}$\\  \textbf{Sayak Ray Chowdhury}$^{3}$\\\\
    \textsuperscript{1}IIT Bombay;
        \textsuperscript{2}UMD College Park; \textsuperscript{3}IIT Kanpur
        \textsuperscript{4}Adobe Research
       }

\begin{document}
\maketitle
\begin{abstract}
In-context learning (ICL) has emerged as a powerful paradigm for adapting large language models (LLMs) to new, data-scarce tasks using only a few carefully selected task-specific examples embedded in the prompt. However, given the limited context size of LLMs, a fundamental question arises: Which examples should be selected to maximize performance on a given user query? While nearest-neighbor-based methods have been widely adopted for this purpose, they suffer from well-known drawbacks in high-dimensional embedding spaces, including poor generalization and limited diversity. In this work, we study example selection in ICL from a principled, information-theoretic perspective. We frame the example selection task as a query-specific high-dimensional optimization problem: selecting a subset of exemplars from a larger example bank to minimize prediction error for a given query. To solve this, we first derive a principled surrogate objective that is approximately submodular, enabling the use of a greedy algorithm with an approximation guarantee. We further enhance our method by introducing an optimal design-based regularizer to encourage diversity in the selected examples. Empirically, we demonstrate significant improvements over standard retrieval methods across a suite of classification tasks, highlighting the benefits of structure-aware, diverse example selection for ICL in real-world, label-scarce scenarios. 
\end{abstract}

\section{Introduction}


With the advent of highly capable large language models (LLMs) ~\citep{adiwardana2020towards, wang2019superglue, zhang2021explain, wang2022survey}, in-context learning (ICL) \citep{rubin2022learning, liu2022makes, wu2022self} via prompt optimization has emerged as a powerful technique for generating responses to complex user queries in data-scarce settings. In this popular and practical paradigm, we assume access to a small bank of high-quality task-specific examples. For a given user query, a few unique and relevant demonstrations are selected to form additional context for the language model. Since LLMs are pre-trained on large corpora, even a small number of carefully chosen exemplars can often suffice to guide the model toward accurate, task-consistent responses \citep{luo2024context}. 
Given the limited context window of large language models, a natural and important question arises for ICL: 
\textit{“Given a specific user query, how can we optimally select and order a subset of task-specific examples from an associated example bank to include in the prompt to maximize performance?”} 
While several existing methods address this question empirically \citep{dong2022survey, luo2024context}, we take a principled, information-theoretic approach.  

Motivated by data-scarce settings and the need for task generalization, we focus on common and practical scenarios where the exemplar retriever is frozen and non-trainable. Among unsupervised retrieval methods, the most widely used is KATE \citep{liu2021makes}, which pioneered a nearest neighbor (kNN)-based strategy for selecting in-context examples. Such methods retrieve examples most similar to the query using pretrained embeddings.
However, as with classical kNN methods, similarity-based retrieval in high-dimensional embedding spaces inherits well-known limitations -- pairwise distances can become less discriminative and more sensitive to noise \citep{beyer1999nearest}. As a result, nearest neighbors may not reliably correspond to the most informative examples for prediction. In machine learning, this problem is often addressed by assuming a structured hypothesis class (e.g., linear models), thereby reducing model complexity and improving generalization guarantees with fewer samples. This motivates our central theoretical question in the ICL problem: \emph{“Can we design an example selection algorithm—i.e., a retriever—that, given a user query, operates under a structured modeling assumption to improve selection quality?”} By incorporating modeling structure into retrieval, we move beyond purely distance-based heuristics and instead select examples that are informative for prediction as well as promote diversity among the selected set.

From a theoretical standpoint, we analyze exemplar selection within a predictive framework in a reproducing kernel Hilbert space (RKHS).\footnote{We do not assume that LLMs are kernel machines; rather, the model serves as a theoretical framework from which we derive a learning-theoretic objective with guarantees and design an algorithm that transfers effectively to practical ICL settings. Many commonly used kernels (e.g. Gaussian RBF, Matern) are universal and their associated RKHS is dense in the space of continuous functions on compact domains \citep{micchelli2006universal}. Hence, RKHS elements are highly expressive and can approximate a wide range of nonlinear functions.}
Under this formulation, we approximate the model responses by a linear function in a high-dimensional (possibly infinite and implicit) feature representation of inputs. The goal of ICL reduces to selecting a small subset of datapoints that minimizes the prediction error for a test query: 
\emph{“Given a test point $\fl{z} \in \mathbb{R}^d$ and an example bank ${(\fl{x}_i, y_i)}_{i=1}^n \in \mathbb{R}^d\times \mathbb{R}$, which $r$ examples should be selected to train a model (predictor) in an RKHS such that the prediction error at $z$ is minimized?”} 
The resulting selection rule naturally induces a retriever for ICL demonstrations in LLMs. Notably, this departs from standard learning-theoretic goals, which aim to generalize across a distribution of test points. In contrast, we focus on query-specific generalization, i.e., training a model that performs well on a single (and potentially arbitrary) test query using a carefully selected subset of training examples. 

To this end, we derive a surrogate loss function over sets that quantifies query-specific prediction error. The negated loss function exhibits two \emph{nice} properties - monotonicity and (approximate) submodularity. This key structural insight allows us to apply a greedy selection algorithm with a provable $1 - e^{-\gamma}$ approximation guarantee relative to the optimal subset, where $\gamma\in (0,1)$ is the approximate sub-modularity ratio of the negated loss with respect to the example bank. In addition to this strategy that selects examples to minimize prediction error, we introduce a mechanism to enforce diversity among the selected examples. Inspired by maximum information gain theory -- a well-established framework in experiment design \citep{lattimore2020bandit} -- we consider the scenario where the goal is to select a subset of examples such that the resulting trained model performs well across all queries in the example bank. We incorporate this design objective as a regularizer into our selection criterion to encourage diversity among selected examples, which improves the model's generalizability and enhances the quality of LLM responses.

Building on this formulation, we design a \emph{computationally efficient and fully unsupervised} example selection algorithm, $\textsc{Kite}$. We focus on the setting where the \emph{retriever is frozen and training-free}. In many real-world deployments, labeled data is scarce, models are accessed via APIs, and retraining/fine-tuning a retriever is either costly or infeasible. Classic top-$r$ similarity-based retrieval methods are widely used precisely because they are model-agnostic, task-agnostic, and deployable off-the-shelf. Our work falls under this setting, and our goal is to provide a \emph{principled theoretical foundation} for exemplar selection that retains these practical advantages while improving performance.\footnote{When training is permissible, our framework can naturally incorporate learned embeddings.} 

We show that \textsc{Kite} preserves the simplicity and negligible overhead of top-$r$ retrieval while delivering consistent gains across model families, dataset characteristics, and embedding encoders. On five classification benchmarks with three open-source LLMs (GPT-Neo-2.7B, Qwen-2.5-1.5B, Llama-3.2-3B), \textsc{Kite} attains the highest accuracy in $13$ of $15$ settings, outperforming kNN-based retrieval~\citep{liu2021makes}, the DPP-based CEIL retriever~\citep{ye2023ceil}. These gains extend to four frontier LLMs (GPT-4o-mini, GPT-4.1, Claude-Haiku-4.5, GPT-5.2) across seven datasets, where \textsc{Kite} achieves the best average on every model, and they remain robust across multiple encoders. 
Beyond classification tasks (which is our theoretical focus), \textsc{Kite} achieves the highest average accuracy across four open-domain QA and reasoning benchmarks on all four frontier models, with gains of $+0.2$ to $+1.8$ points over the best KNN retriever, demonstrating \textsc{Kite}'s generalization.

\subsection{Related Work}

\label{sec:related_works}

\noindent\textbf{In-Context Learning (ICL)} introduced by \citet{brown2020language} is a powerful paradigm in which LLMs learn new tasks by conditioning on a few input-output examples provided in the prompt, without any parameter updates. The underlying mechanisms of ICL have been extensively studied; some works suggest that ICL allows the model to infer a shared latent concept from demonstrations \citep{xie2021explanation}. Further research indicates that ICL is a nuanced process, as models do not always rely strictly on the provided input-output mappings \citep{min2022rethinking}. A broad range of studies has also investigated the factors that influence ICL performance, such as demonstration calibration and corpora effects \citep{zhao2021calibrate, shin2022effect, wei2022emergent, yoo2022ground, wei2023larger}.

\vspace{0.3cm}
\noindent\textbf{Example retrieval for few-shot learning.} The effectiveness of ICL is highly sensitive to the choice of examples, leading to extensive work on example selection \citep{brown2020language, min2022rethinking}. Prior methods can be broadly categorized into training-free and training-based retrieval strategies.

Training-free (unsupervised) retrieval methods select examples at inference time without updating retriever parameters. Classic approaches use semantic similarity (e.g., nearest neighbors in embedding space) to choose top-$r$ exemplars \citep{liu2022makes}. Diversity-enhancing subset selection strategies, most notably CEIL~\citep{ye2023ceil} -- which uses a determinantal point process (DPP) score over a kernel similarity matrix to encourage diverse exemplar selection -- also fall into this category. \textsc{Kite} also operates in this regime but differs in that its selection objective is derived from a principled kernelized predictive framework with provable approximation guarantees.

In contrast, training-based methods learn a retriever to select effective demonstrations. These approaches often fine-tune dense retrievers using labels distilled from a language model~\citep{rubin2022learning}, train neural networks to predict ICL performance~\citep{wu2024prompt}, or employ contrastive objectives to balance relevance and coverage \citep{li2023unified, nguyen2023context, luo2023dr, wang2023latent, ye2023compositional, ghosal2025relic, ghosal2025promptrefine}. Such methods improve task adaptation but require labeled data and retriever training, which introduce additional deployment complexity.

\section{Problem Formulation}
\label{sec:problem-statement}

Consider access to a dataset $\mathcal{X}$ of $n$ known input--output pairs $(\fl{x}_1,y_1),\ldots,(\fl{x}_n,y_n)$, where inputs are $d$-dimensional vectors and outputs are real scalars. Given a test query $\mathbf{z} \in \mathbb{R}^d$ and $r \ll n$, the in-context learning (ICL) problem is to select an optimal subset of $r$ examples from $\mathcal{X}$ as context for the LLM's prediction on $\mathbf{z}$, so as to minimize prediction error. The selected subset must contain unique elements, since duplicating examples does not provide additional context in LLMs.

We assume that, conditioned on $\fl{z}$, the model response $y$ for any input $\fl{x}$ is a noisy evaluation of an unknown function $h$ from a function class $\mathcal{H}$, i.e., $y = h(\fl{x}) + \eta$, where $\eta$ is zero-mean sub-Gaussian noise. Although LLM outputs are token sequences, we treat $y$ as a scalar task-relevant signal extracted from the output --- for classification tasks, $y$ is the likelihood of the correct label given $\fl{x}$. This scalar regression view is standard in the ICL theory literature \citep{xie2021explanation,garg2022can}, and the noise term absorbs stochasticity from sampling, prompt-order effects, and tokenizer-level variability. The function $h$ is defined with respect to the test query $\fl{z}$, and the selected subset $\ca{S}$ of $r$ examples is used to train the model to predict $\fl{z}$.

We focus on ICL tasks with single-valued outputs, such as classification and regression. In these settings, the ICL problem reduces to: \emph{for a given test query $\fl{z}$, which subset $\ca{S}$ of size at most $r$ from $\mathcal{X}$ should we select so that a predictor $\widehat{h}_{\ca{S}} \in \mathcal{H}$ fitted on $\ca{S}$ minimizes the expected error at $\fl{z}$?} Formally,
\begin{equation}\label{eq:disc}
\argmin_{\ca{S} \subseteq \mathcal{X} \,:\, |\ca{S}| \le r}
\big| h(\fl{z}) - \widehat{h}_{\ca{S}}(\fl{z}) \big|~.
\end{equation}
This discrete problem is computationally challenging, with complexity exponential in $r$ and polynomial in $n$ due to the $\binom{n}{r}$ possible subsets. Since this runtime would be incurred per query, and fast inference without sacrificing accuracy is paramount, our goal is an efficient algorithm that finds high-quality solutions to~\eqref{eq:disc}. The resulting selection rule can be used directly for ICL by providing the chosen examples as demonstrations; the selected subset thus acts as a proxy for training a predictor in $\mathcal{H}$.

To make the problem tractable, we assume $\ca{H}$ is a reproducing kernel Hilbert space (RKHS, see~\citet{scholkopf2002learning}). An RKHS is fully characterized by a symmetric, positive semi-definite kernel function $k: \bb{R}^d \times \bb{R}^d \to \bb{R}$, and vice versa. Three commonly used kernels are linear: $k_{\text{lin}}(\fl{x},\fl{x}')= \fl{x}^\top \fl{x}'$, polynomial with degree $m$: $k_{\text{Poly}}(\fl{x},\fl{x}')= (\fl{x}^\top \fl{x}'+c)^m$ and Gaussian/RBF with length-scale $\sigma$:
$k_{\text{Gauss}}(\fl{x},\fl{x}')= \exp\left(-  \frac{\| \fl{x}-\fl{x}'\|^2}{2\sigma^2}\right)$.  The kernel trick gives a feature map $\phi : \bb{R}^d \to \ca{H}$ with $k(\fl{x},\fl{x}') = \phi(\fl{x})^\top \phi(\fl{x}')$, where the inner product is that of $\ca{H}$. By the \emph{reproducing property}, $h(x) = \phi(\fl{x})^\top h$ for all $h \in \mathcal{H}$ and $x \in \mathbb{R}^d$. Together, these let us perform all computations in the high-dimensional (possibly infinite) Hilbert space $\ca{H}$ analogously to Euclidean space, without explicitly computing inner products.

\section{Algorithm Design}

We develop a principled two-stage approach to solve the in-context learning problem: reformulate ~\eqref{eq:disc} as a submodular optimization problem and a greedy algorithm to solve it.

\subsection{Problem Reformulation} 

We first invoke the Representer theorem to find a predictor $\widehat h_{\ca{S}} \in \ca{H}$. Consider a fixed, known, positive semi-definite kernel $k$ and the associated (implicit) feature map $\phi$. By the reproducing property, the kernel ridge regressor with a regularizer $\beta > 0$ takes the form~\cite{scholkopf2002learning}
\begin{align*}
    \widehat h_{\ca{S}}(\fl{z})= \phi(\fl{z})^\top \mathbf{V}_\mathcal{S}^{-1}\sum\nolimits_{i \in \mathcal{S}} \phi(\mathbf{x}_i) \, y_{i}~,
\end{align*}
where $\mathbf{V}_\mathcal{S} =\sum_{i \in \mathcal{S}} \phi(\mathbf{x}_i)\phi(\mathbf{x}_i)^\top+ \beta\mathbf{I}$ denotes the $\beta$-regularized design covariance matrix and $\mathbf{I}$ denotes the identity matrix (operator) in $\ca{H}$. Now, applying Chernoff bound for sub-Gaussian random variables, we can bound, for any $\delta \in (0,1)$, the prediction error for the test query $\mathbf{z}$ as

\begin{equation*}\label{eq:conc}
\left| h(\fl{z}) - \widehat{h}_{\ca{S}}(\fl{z})\right| \lessapprox \!\sqrt{\phi(\mathbf{z})^\top \mathbf{V}_{\mathcal{S}}^{-1} \phi(\mathbf{z}) \!\log(1/\delta)}
\end{equation*}
with probability $\ge 1-\delta$, where a noise-variance dependent factor is hidden under $\lessapprox$. This is a well-known result on the prediction error in RKHSs~\citep[Chapter 20]{lattimore2020bandit}. It shows that the upper bound on prediction error is related to the chosen subset $\mathcal{S}$ via the term $\phi(\mathbf{z})^\top \mathbf{V}_{\mathcal{S}}^{-1} \phi(\mathbf{z})$. Hence, we need to minimize this for a fixed $\mathbf{z}\in \mathbb{R}^d$, or, equivalently solve the optimization problem:

\begin{equation}\label{eq:set-func}
\!\!\argmax_{\mathcal{S} \subseteq \mathcal{X},\, |\mathcal{S}|\le r} \!\!f_\mathbf{z}(\mathcal{S}), \, f_\mathbf{z}(\mathcal{S}) = -\phi(\mathbf{z})^\top \mathbf{V}_{\mathcal{S}}^{-1} \phi(\mathbf{z}).
\end{equation}

Note that we can write the set function $f_\mathbf{z}$ in terms of the kernel $k$, so that it can be implemented solely with kernel computations, avoiding matrix inverses and matrix-vector multiplications in (possibly) infinite-dimensional RKHSs. We start with a few notations -
for a set $\mathcal{S} \equiv \{\fl{x}_{i_1},\ldots,\fl{x}_{i_{|\mathcal S|}}\}$, define a matrix $\Phi_{\mathcal S}$ by stacking the feature vectors $\phi(\fl{x}_{i_1}),\ldots,\phi(\fl{x}_{i_{|\mathcal S|}})$
row-wise.

Next, define the kernel vector between an input $x$ and the set $\mathcal S$ as $\mathbf{k}_{\mathcal S}(x)
=\Phi_{\mathcal S}\phi(\fl{x})$.

The corresponding kernel (Gram) matrix admits
$\mathbf{K}_{\mathcal S}
=
\Phi_{\mathcal S}\Phi_{\mathcal S}^{\top}
$ and the regularized design covariance matrix becomes $\mathbf{V}_{\mathcal S} = \Phi_{\mathcal S}^{\top}\Phi_{\mathcal S}+\beta \mathbf{I}$. 
Then, using Woodbury identity together with the kernel trick, we obtain
\begin{align}\label{eq:kernel-trick}
&f_{\fl{z}}(\ca{S})= -\frac{1}{\beta}\cdot k_{\ca{S}}(\fl{z},\fl{z})~,\text{where}\\
&k_{\ca{S}}(\fl{x},\fl{y})\!=\!k(\fl{x},\fl{y}) \!-\! \mathbf{k}_{\mathcal S}(\fl{x})^{\top}
\!(\mathbf{K}_{\mathcal S}\!+\!\beta \mathbf{I}_{|\ca{S}|})^{-1}
\mathbf{k}_{\mathcal S}(\fl{y})\nonumber   
\end{align}
denotes the similarity between inputs $\fl{x}$ and $\fl{y}$ conditioned on the set $\ca{S}$ (derivation in appendix).

The set function $f_\mathbf{z}$ is monotonically increasing, since for any $\mathcal{S} \subseteq \mathcal{L}$, the design covariance matrices satisfy $\fl{V}_{\ca{S}} \preceq \fl{V}_{\ca{L}}$ in the Lowener order. Along with monotonicity, a desirable property for maximizing set functions is submodularity, defined below.
\begin{definition}[Submodular function]
A set function \( f: 2^{\ca{X}} \rightarrow \mathbb{R} \) is called \emph{submodular} if for all \( \ca{S} \subseteq \ca{L} \subseteq \ca{X} \) and \( \fl{x} \in \ca{X} \setminus \ca{L} \), it satisfies
\[
f(\ca{S} \cup \{\fl{x}\}) - f(\ca{S}) \geq f(\ca{L} \cup \{\fl{x}\}) - f(\ca{L})~.
\]
\end{definition}
\noindent For monotone and submodular functions, a greedy algorithm admits an $(1-\nicefrac{1}{e})$-factor approximation. However, $f_{\fl{z}}$ isn't submodular. Instead, it exhibits approximate submodularity and retains the near-optimality guarantee of greedy algorithms. This approximation is captured by the \emph{submodularity ratio} \cite{das2011submodular}, defined below.

\begin{definition}[Submodularity ratio]
\label{def:submod_ratio}
For a ground set \( \ca{X} \) and a parameter \( r \in \mathbb{N} \), the submodularity ratio of a set function \( f: 2^{\ca{X}} \rightarrow \mathbb{R} \) is defined as
\[
\gamma_r(f) = \!\!\min_{\substack{\ca{S},\ca{L} \subseteq \ca{X}: |\ca{L}| \leq r,\\ \ca{L} \cap \ca{S} = \emptyset}} 
\!\!\!\frac{\sum_{\fl{x} \in \ca{L}} \left( f(\ca{S} \cup \{\fl{x}\}) - f(\ca{S}) \right)}{f(\ca{S} \cup \ca{L}) - f(\ca{S})}.
\]
\end{definition}
\noindent The submodularity ratio captures how much the value of $f$ can increase by adding any subset $\ca{L}$ of size at most $r$ to $\ca{S}$, compared to the combined benefits of adding its elements to $\ca{S}$. It quantifies the degree to which \( f \) satisfies the diminishing returns property, and hence how well greedy algorithms perform in maximizing \( f \) under cardinality constraints. If \( f \) is submodular, then \( \gamma_r = 1 \). When \(\gamma_r < 1 \), the function is approximately submodular.

\begin{lemma}[Submodularity ratio of $f_{\fl{z}}$]\label{lem:submod}
    For any $\fl{z} \in \mathbb{R}^d$ and $\beta > 0$, 
$\gamma_r(f_{\fl{z}}) \ge \frac{1}{1+(r-1)\mu}$,
    where $\mu= \max\limits_{\fl{x}_i,\fl{x}_j \notin \ca{S}}\frac{|k_{\ca{S}}(\fl{x}_i,\fl{x}_j)|}{\sqrt{\beta+k_{\ca{S}}(\fl{x}_i,\fl{x}_i)\sqrt{\beta+k_{\ca{S}}(\fl{x}_j,\fl{x}_j)}}}$ is the maximum coherence between any pair in the set $\ca{X}\setminus \ca{S}$.
\end{lemma}

\subsection{Greedy Algorithm}

\begin{algorithm*}[t]
\caption{\textsc{Kernelized Information Theoretic Exemplars} (\textsc{Kite})}
\label{alg:submodular_icl}
\begin{algorithmic}[1]

\Require Dataset $\mathcal{X}=\{(\mathbf{x}_{i},y_{i})\}_{i=1}^n \in \mathbb{R}^d\times \mathbb{R}$; kernel function $k$; test query $\mathbf{z}$; subset size $r$; regularization $\beta>0$; diversity weight $\lambda>0$

\State Initialize available datapoints $\mathbf{mask} \gets \mathbf{1}_n$, selected set $\mathcal{S} \gets \emptyset$, kernel $k_{\ca{S}}(\fl{z},\fl{x}) \gets k(\fl{z},\fl{x})$
\For{$i = 1$ to $r$}
    \State $\mathcal{X}_{\text{available}} \gets \{\mathbf{x}_{j} : \mathbf{mask}_j = 1\}$
    \For{each $\mathbf{x} \in \mathcal{X}_{\text{available}}$}
  
        \State $\mathrm{rel}(\mathbf{x}) \gets \frac{\left(k_{\ca{S}}(\fl{z},\fl{x})\right)^2}{\beta + k_{\ca{S}}(\fl{x},\fl{x})}$, 
        where $k_\ca{S}(\fl{z},\fl{x})=k(\fl{z},\fl{x})-\fl{k}_{\ca{S}}(\fl{z})^{\top}(\fl{K}_\ca{S}+\beta \fl{I}_{|\ca{S}|})^{-1}\fl{k}_\ca{S}(\fl{x})$ 
 
        \State       $\mathrm{div}(\mathbf{x}) \gets \log \big(\beta + k_{\ca{S}}(\fl{x},\fl{x})\big)$
        \State $\mathrm{score}(\mathbf{x}) \gets \mathrm{rel}(\mathbf{x}) + \lambda \cdot \mathrm{div}(\mathbf{x})$
    \EndFor
    \State $\mathbf{x}^* \gets \operatorname*{argmax}_{\mathbf{x} \in \mathcal{X}_{\text{available}}}\, \mathrm{score}(\mathbf{x})$
    \State $\mathcal{S} \gets \mathcal{S} \cup \{\mathbf{x}^*\}$, update $\mathbf{mask}$ to exclude $\mathbf{x}^*$
\EndFor
\State \Return $\mathcal{S}$
\end{algorithmic}
\end{algorithm*}


\noindent\textbf{Selecting relevant examples.} For any $\mathcal{S'} \subset \mathcal{S}$ such that $\mathcal{S} \setminus \mathcal{S'} = \{\mathbf{x}\}$, using Sherman-Morrison formula and kernel trick (\eqref{eq:kernel-trick}), we get
\begin{align*}
f_\mathbf{z}(\mathcal{S}) 
= f_\mathbf{z}(\mathcal{S'}) + \frac{1}{\beta} \cdot \frac{\left(k_{\ca{S'}}(\fl{z},\fl{x})\right)^2}{\beta+ k_{\ca{S'}}(\fl{x},\fl{x})}
~.
\end{align*}
This naturally yields a greedy algorithm. Starting with an empty set $\ca{S}_0 = \emptyset$, at each step $i \in \lbrace 1,2,\ldots,r\rbrace$, select the 
example that provides the maximum marginal gain over already chosen examples $\ca{S}_{i-1}=\lbrace \fl{x}_1,\ldots,\fl{x}_{i-1} \rbrace$, i.e., select
\begin{equation}\label{eq:rel-greedy}
\mathbf{x}_i = \operatorname*{argmax}_{\mathbf{x} \in \mathcal{X} \setminus \mathcal{S}_{i-1}}\frac{\left(k_{\ca{S}_{i-1}}(\fl{z},\fl{x})\right)^2}{\beta+ k_{\ca{S}_{i-1}}(\fl{x},\fl{x})} ~.
\end{equation}

Adapting from \citet{das2011submodular}, we obtain the following guarantee for the greedy algorithm.

\begin{theorem}[Greedy algorithm performance]\label{thm:greedy}
For any $\fl{z} \in \mathbb{R}^d$, let the greedy algorithm return a set \( \hat{\ca{S}} \) for the optimization problem in \eqref{eq:set-func}. Then
$f_{\fl{z}}(\hat{\ca{S}}) \geq \left(1 - e^{-\gamma_r(f_\fl{z})} \right) \cdot f(\ca{S}^\ast)$,
where \( S^\ast \) is an optimal solution of size at most \( r \).
\end{theorem}
\noindent
The approximation guarantee degrades gracefully with the submodularity ratio \( \gamma_r(f_{\fl{z}}) \). When \( \gamma_r(f_{\fl{z}}) \approx 1 \), the greedy algorithm performs nearly as well as the optimal algorithm (up to an $(1-1/e)$ factor). Increasing the regularization strength \( \beta \) drives \( \gamma_r(f_{\fl{z}}) \to 1 \). In practice, \( \gamma \) can be estimated, providing useful certificates of near-optimality for greedy selection. Empirical verification on the datasets used in this work (Appendix~\ref{app:submodularity}, Table~\ref{tab:gamma}) shows that the realized submodularity ratio is typically in the $0.7$--$0.9$ range and approaches $1$ as $r$ or $\beta$ grow; the lowest values are observed at small $r$ on HellaSwag (down to $\sim\!0.4$ at $r{=}5$), where greedy selection still performs well empirically (Table~\ref{tab:classification-results}).

\noindent\textbf{Selecting diverse examples.} Equation~\ref{eq:rel-greedy} seeks the example $\fl{x}_i$ which is most relevant to the input query $\fl{z}$ in the geometry induced by the kernel function $k_{\ca{S}_{i-1}}(\cdot,\cdot)$.
However, for ICL, both relevance (i.e., choosing examples similar to the input) and diversity
(i.e., choosing diverse examples) are essential, and need to be carefully balanced.
To select diverse examples, we seek the set $\ca{S}$ that fetches the maximum information about the unknown function $h \in \ca{H}$ from noisy responses. If we assume $h$ to be sampled from a Gaussian process\footnote{A Gaussian process (GP) over $\mathbb{R}^d$, is a collection of random variables $(h(x))_{x \in \mathbb{R}^d}$, such that every finite sub-collection $(h(x_i))_{i = 1}^m$ is jointly Gaussian~\cite{williams2006gaussian}. The GP model only aids in designing the diversity objective, and has nothing to do with the actual function $h$, which is a fixed, unknown, member of the RKHS $\ca{H}$.} with covariance function $k$, then the information gain about $h$ from $r$ noisy responses is given by $g(\ca{S})=\log \det \left(  \fl{K}_{\mathcal{S}}+\beta\fl{I}_r\right)$, where $\fl{K}_{\mathcal{S}}$ is the Gram matrix over the set $\ca{S}$~\citep{cover1999elements}.

The set function $g$ is known as the D-optimal design~\citep{pukelsheim2006optimal} and the most informative (i.e., diverse) set is the one that maximizes $g$.

\begin{lemma}[Submodularity]\label{lem:submod_div}
    For any $\beta>0$ and $r \in \mathbb{N}$, $g(\ca{S})= \log \det(\fl{K}_\mathcal{S}+\beta \fl{I}_r)$ is monotone, submodular, and admits the greedy selection
    \begin{equation}\label{eq:div-greedy}
\mathbf{x}_i = \operatorname*{argmax}_{\mathbf{x} \in \mathcal{X} \setminus \mathcal{S}_{i-1}} \log\big(\beta + k_{\ca{S}_{i-1}}(\fl{x},\fl{x})\big)~.
\end{equation}
\end{lemma}
\noindent\textbf{Combining relevance and diversity.}
To select relevant as well as diverse examples (set-conditioned), we combine 
\eqref{eq:rel-greedy} and \eqref{eq:div-greedy} to obtain
\begin{align}\label{eq:combine}
\fl{x}_i = \operatorname*{argmax}_{\fl{x}\in \ca{X}\setminus \ca{S}_{i-1}}
&\,\frac{\big(k_{\ca{S}_{i-1}}(\fl{z},\fl{x})\big)^2}{\beta + k_{\ca{S}_{i-1}}(\fl{x},\fl{x})}\\
&+ \lambda \log \big(\beta + k_{\ca{S}_{i-1}}(\fl{x},\fl{x})\big)~,\nonumber\
\end{align}
where $\lambda \geq 0$ controls the relevance-diversity trade-off. The combined objective remains monotone and approximately submodular, so the set $\hat{\ca{S}}$ obtained from \eqref{eq:combine} achieves a guarantee similar to Theorem~\ref{thm:greedy}. We call this algorithm \textbf{K}ernelized \textbf{I}nformation \textbf{T}heoretic \textbf{E}xemplars (\textsc{Kite}) for ICL; pseudo-code is in Algorithm~\ref{alg:submodular_icl}. In our experiments, the kernel is a tunable hyperparameter of \textsc{Kite}.

\section{Experiments}
\label{sec:exp}
\vspace{-2mm}

We present a comprehensive empirical analysis of \textsc{Kite} along multiple axes: (i)~base accuracy on five classification benchmarks across three open-source LLMs (Table~\ref{tab:classification-results}); (ii)~scaling to 7B/8B variants (Table~\ref{tab:scaling-sst2-cmsqa}); (iii)~generalization to four frontier API-only LLMs (GPT-4o-mini, GPT-4.1, Claude-Haiku-4.5, GPT-5.2) on seven classification datasets (Tables~\ref{tab:frontier_summary},~\ref{tab:frontier_perdataset}); (iv)~kernel ablations on both open-source and frontier settings (Tables~\ref{tab:kernel-ablation},~\ref{tab:kernel_frontier}); (v)~robustness across multiple encoders (Tables~\ref{tab:embedding-ablation},~\ref{tab:nvembed_summary}); (vi)~sensitivity to the exemplar budget $r$ (Tables~\ref{tab:k10_summary},~\ref{tab:k5vsk10}; Fig.~\ref{fig:ice_ablation}); (vii)~empirical verification of the submodularity property (App.~\ref{app:submodularity}, Table~\ref{tab:gamma}); and (viii)~wall-clock runtime analysis showing negligible overhead vs.\ standard retrieval (App.~\ref{app:runtime}).

\begin{table*}[t]
\centering
\footnotesize
\setlength{\tabcolsep}{2pt}
\renewcommand{\arraystretch}{0.9} 
\resizebox{0.99\textwidth}{!}{
\begin{tabular}{l|ccc|ccc|ccc|ccc|ccc|ccc}
\toprule
& \multicolumn{3}{c|}{\textbf{SST-2}} 
& \multicolumn{3}{c|}{\textbf{SST-5}} 
& \multicolumn{3}{c|}{\textbf{CMSQA}} 
& \multicolumn{3}{c|}{\textbf{MRPC}} 
& \multicolumn{3}{c|}{\textbf{QNLI}}
& \multicolumn{3}{c}{\textbf{HellaSwag}} \\
\cmidrule(lr){2-4} \cmidrule(lr){5-7} \cmidrule(lr){8-10} \cmidrule(lr){11-13} \cmidrule(lr){14-16} \cmidrule(lr){17-19}
\textbf{Method} 
  & GN & QW & LL 
  & GN & QW & LL 
  & GN & QW & LL 
  & GN & QW & LL 
  & GN & QW & LL
  & GN & QW & LL \\
\midrule
Random & 63.36 & 86.32 & 89.84 
       & 32.31 & 40.53 & 42.41 
       & 42.25 & 65.90 & 67.22 
       & 66.38 & 70.45 & 71.12 
       & 57.56 & 70.21 & 68.43 
       & 41.75 & 66.30 & 65.21     \\
BM25   & 67.78 & 90.14 & 91.63
       & 36.05 & 47.86 & \textbf{47.85} 
       & 42.91 & 70.02 & 70.92 
       & 66.96 & 73.77 & 70.58 
       & 62.12 & \textbf{71.11} & 69.38 
       & 41.34  & 67.85  & 66.32     \\

Dense  & 72.94 & 88.99 & 92.55
       & 35.96 & 46.64 & 44.14 
       & 43.98 & 70.18 & 72.23 
       & 65.83 & 73.92 & 71.81 
       & 64.42 & 70.73 & 70.04 
       & 40.44 & 68.80 & 67.88     \\

CEIL   & 71.22 & 90.25  & 91.86
       & 36.84  & 47.88  & 46.45
       & 37.53  & 70.05  & 71.65
       & 68.04  & 74.25  & 70.63
       & 63.62  & 70.53  & 70.08
       & 40.56  & 69.32  & 68.55 \\

\textsc{Kite}   & \textbf{74.41}  & \textbf{93.35}  & \textbf{94.28}  
       & \textbf{40.60} & \textbf{49.59} & 47.59 
       & \textbf{46.60} & \textbf{71.25} & \textbf{72.89} 
       & \textbf{68.38} & \textbf{75.27} & \textbf{71.98} 
       & \textbf{65.32} & 71.05 & \textbf{71.68} 
       & \textbf{41.68}     & \textbf{71.02}  & \textbf{70.12}\\

\bottomrule
\end{tabular}
}
\vskip -2mm
\caption{\textbf{Evaluation Results.} We compare classification accuracy (\%) of \textsc{KITE} against retrieval baselines on six few‐shot benchmarks. Model abbreviations: GN (GPT-Neo 2.7B), QW (Qwen 2.5–1.5B), LL (Llama 3.2–3B). }
\label{tab:classification-results}
\end{table*}

\begin{table*}[t]
\centering
\footnotesize
\setlength{\tabcolsep}{4pt}
\renewcommand{\arraystretch}{0.82}
\resizebox{0.72\textwidth}{!}{%
\begin{tabular}{lcccccccc}
\toprule
& \multicolumn{4}{c}{\textbf{SST-2}}
& \multicolumn{4}{c}{\textbf{CMSQA}} \\
\cmidrule(lr){2-5} \cmidrule(lr){6-9}
\textbf{Method}
& \textbf{QW-1.5B} & \textbf{QW-7B} & \textbf{LL-3B} & \textbf{LL-8B}
& \textbf{QW-1.5B} & \textbf{QW-7B} & \textbf{LL-3B} & \textbf{LL-8B} \\
\midrule
Zero-shot & 61.69 & 65.82 & 68.73 & 74.21 & 57.00 & 64.53 & 58.86 & 61.52 \\
Random    & 63.36 & 71.53 & 89.84 & 90.12 & 65.90 & 68.29 & 67.22 & 68.95 \\
Dense     & 88.99 & 89.45 & 92.55 & 93.35 & 70.18 & 72.35 & 72.23 & 74.19 \\
CEIL      & 90.25 & 91.63 & 91.86 & 93.61 & 70.05 & 74.82 & 71.65 & 73.83 \\
\textsc{Kite}
           & \textbf{93.35} & \textbf{94.50} & \textbf{94.28} & \textbf{94.84}
           & \textbf{71.25} & \textbf{75.69} & \textbf{72.89} & \textbf{75.22} \\
\bottomrule
\end{tabular}%
}
\vskip -2mm
\caption{\textbf{Scaling results.} Accuracy (\%) when scaling from smaller backbones to larger ones. QW denotes Qwen-2.5, and LL denotes Llama-3.2. \textsc{Kite} consistently outperforms zero-shot and retrieval-based baselines across model sizes.}
\label{tab:scaling-sst2-cmsqa}
\vspace{-3mm}
\end{table*}

\vspace{0.3cm}
\noindent\textbf{Implementation details.} For experiments with open-source LLM, we use $\beta\!=\!0.02,\lambda\!=\!0.5$, and \texttt{bert-base-uncased} \citep{devlin2019bertpretrainingdeepbidirectional} embeddings for retrieval. Our main open-source experiments (Table~\ref{tab:classification-results}) use three LLMs: GPT-Neo-2.7B~\citep{Black2021GPTNeoLS}, Qwen-2.5-1.5B~\citep{hui2024qwen25codertechnicalreport}, and Llama-3.2-3B~\citep{grattafiori2024llama3herdmodels}, with $r{=}50$ in-context examples (truncated to fit the context window) and three kernels: Linear, Polynomial (degree $m{=}3$), and Gaussian/RBF ($\sigma{=}1.0$). 

For experiments with Frontier-LLM, we use $r{=}5,10$ (matching realistic API usage) and an expanded kernel suite (Linear, RBF, Laplacian, Mat\'{e}rn-3/2, Rational Quadratic) - Tables ~\ref{tab:frontier_perdataset}, \ref{tab:k10_perdataset}, \ref{tab:k10_perdataset_other} and \ref{tab:nvembed_perdataset_a}. 
We use two distinct encoders - OpenAI's \texttt{text-embedding-3-large} and NVIDIA's \texttt{NV-Embed-v2} embeddings of 3072 and 4096 dimensions respectively.
For each frontier experiment we use $500$ inference queries and a candidate bank of $2000$ examples (first from val/train splits respectively). In Tables~\ref{tab:frontier_summary} and~\ref{tab:frontier_perdataset}, \textsc{Kite} is reported with $\lambda$ fixed at $0.5$ and the best accuracy across the five kernels; per-kernel and per-$\lambda$ breakdowns appear in Tables~\ref{tab:kernel-ablation},~\ref{tab:kernel_frontier}, and App.~\ref{app:kernel_guidance}.
For reproducibility, we follow a uniform protocol: embeddings are used directly from the pretrained encoder with no centering, whitening, or per-dataset re-scaling; the kernel hyperparameters above ($\sigma$, $m$, polynomial constant $c{=}1$, Laplacian/Mat\'{e}rn/RQ length-scales $=\!1$) and the model hyperparameters $\beta$ and $\lambda$ are \emph{fixed across all datasets and models}. The same fixed values are used for all baselines wherever they admit a kernel parameter, ensuring the comparison is not biased toward \textsc{Kite}.

\vspace{0.3cm}
\noindent\textbf{Datasets.} 
A detailed description of the datasets is given in Section \ref{sec:dataset-details}. For the open-source experiments, to empirically demonstrate the efficacy of \textsc{Kite}, we evaluate on five few-shot classification datasets: SST-2 \& 5 \citep{socher-etal-2013-recursive} (single-sentence sentiment), CMSQA \citep{talmor-etal-2019-commonsenseqa} (question answering), MRPC \citep{dolan-etal-2004-unsupervised} (sentence-pair paraphrase), QNLI \citep{wang-etal-2018-glue} (binary entailment classification), and HellaSwag \citep{zellers-etal-2019-hellaswag} (common sense NLI). For each task, we use the corresponding validation split for evaluation and employ the deduplicated training split as the candidate exemplar set.

\vspace{0.3cm}
\noindent\textbf{Evaluation metrics.}
Following \citet{brown2020languagemodelsfewshotlearners}, we formulate all classification tasks as multiple-choice problems. Input prompts concatenate the context with each candidate label. We compute the conditional log-likelihood of the label tokens given the prompt, and select the candidate with the highest log-likelihood as the prediction. Performance is evaluated using accuracy on the validation split. For all methods, we use the retriever's own ranking as the prompt order: kNN uses descending similarity, CEIL uses its greedy DPP-selection order, and \textsc{Kite} uses its greedy acquisition order. This ensures that observed differences are attributable to \emph{which} exemplars are selected rather than \emph{how} they are ordered.
For frontier models accessed via API, logits are unavailable; we instead prompt each model to generate the answer directly and report accuracy as exact match of the generated token to the gold label.

\vspace{0.3cm}
\noindent\textbf{Baselines.} We compare \textsc{Kite} against Random, BM25 \citep{robertson2009probabilistic}, top-$r$ with Dense embeddings~\citep{liu2021makes}, and the training-free version of CEIL~\citep{ye2023ceil}, a DPP-based retrieval strategy.


\vspace{0.3cm}
\noindent\textbf{Results.} We report evaluation results in Table~\ref{tab:classification-results}. We find that \textsc{Kite} consistently outperforms all baselines across most datasets and model architectures. On GPT-Neo-2.7B, for example, \textsc{Kite} delivers an accuracy improvement of +$4.55\%$ on SST-5~\citep{socher-etal-2013-recursive} and +$3.69\%$ on CMSQA~\citep{talmor-etal-2019-commonsenseqa} over BM25~\citep{robertson2009probabilistic}. This trend is consistently observed across all evaluated models. With Qwen-2.5-1.5B, \textsc{Kite} surpasses the strongest baseline, CEIL~\citep{ye2023ceil}, on four out of five datasets, achieving notable gains of +$1.71\%$ on SST-5 and +$1.70\%$ on HellaSwag. On Llama-3B, \textsc{Kite} achieves the highest accuracy on four datasets, including a +$2.24\%$ improvement on HellaSwag over CEIL~\citep{ye2023ceil}. Across all $15$ evaluation settings (five datasets and three models), \textsc{Kite} attains the highest accuracy in $13$ cases, underscoring its robustness and efficacy as an exemplar retrieval framework.

\vspace{0.3cm}
\noindent\textbf{Computational cost.} A natural concern is whether \textsc{Kite}'s greedy selection imposes meaningful overhead at $r{=}50$. We measure end-to-end wall-clock time per query on SST-5 ($8534$ examples) using Qwen-2.5-1.5B/7B on a single A6000 GPU. \textsc{Kite}'s selection adds only $\le 0.03$\,s per query on the 1.5B model and $\le 0.19$\,s on the 7B relative to BM25, leaving end-to-end latency effectively unchanged (App.~\ref{app:runtime}, Table~\ref{tab:runtime}).

\vspace{0.3cm}
\noindent\textbf{Scaling to larger models.} To study the effect of model scale, we evaluate KITE on SST-2~\citep{socher-etal-2013-recursive} and CMSQA~\citep{talmor-etal-2019-commonsenseqa} using both smaller and larger variants of Qwen-2.5 and Llama-3.2. As shown in Table~\ref{tab:scaling-sst2-cmsqa}, KITE consistently outperforms all baselines across model sizes, and the gains persist as we scale to larger models such as Qwen-2.5-7B and Llama-3.2-8B.

\vspace{0.3cm}
\noindent\textbf{Scaling to Frontier LLMs.}\label{sec:frontier} To validate that \textsc{Kite}'s gains extend beyond open-source models with accessible logits, we evaluate on four frontier LLMs accessed via API: GPT-4o-mini, GPT-4.1, Claude-Haiku-4.5, and GPT-5.2. We use OpenAI's \texttt{text-embedding-3-large} for retrieval (as the deployment-realistic choice in an API setting), with $r{=}5$, $\beta{=}0.02$, $\lambda{=}0.5$, and evaluate on seven classification datasets: SST-5, MRPC, QNLI, SWAG~\citep{zellers2018swag}, IMDB, DBPedia, and BoolQ, reporting the best result across five kernels. Table~\ref{tab:frontier_summary} presents the average accuracy.
\textsc{Kite} achieves the highest average accuracy on all four models, with gains of $+0.8$ to $+1.3$ points over the best KNN retriever and $+0.7$ to $+1.2$ over CEIL. The improvements are consistent across capability tiers, from the lightweight GPT-4o-mini to the more capable GPT-5.2, showing that \textsc{Kite}'s information-theoretic selection provides complementary benefits to model capability.
 
\begin{table}[t]
\centering
\small
\resizebox{\linewidth}{!}{%
\begin{tabular}{lccccc}
\toprule
\textbf{Model} & \textbf{BM25} & \textbf{KNN} & \textbf{CEIL} & \textbf{\textsc{Kite}} \\
\midrule
GPT-4o-mini       & 79.0 & 80.5 & 80.4 & \textbf{81.3} \\
GPT-4.1           & 80.7 & 81.2 & 81.2 & \textbf{82.2}  \\
Claude-Haiku-4.5  & 81.1 & 82.3 & 82.8 & \textbf{83.5}  \\
GPT-5.2           & 79.7 & 81.6 & 81.7 & \textbf{82.9}\\
\bottomrule
\end{tabular}}
\vskip -2mm
\caption{Average classification accuracy (\%) on seven classification datasets with frontier LLMs ($r{=}5$, OpenAI large text embeddings). For each method, we report the best result across five kernels. \textsc{Kite} achieves the highest accuracy on every model.}
\label{tab:frontier_summary}
\vspace{-0.5cm}
\end{table}

\vspace{0.3cm}
\noindent\textbf{Ablation study on kernel function.}
The kernel is a critical hyperparameter in \textsc{Kite}, as it defines the geometry of the feature space. 
We conduct an ablation comparing three kernels: {Linear}, {Polynomial} (degree $m=3$), and {Gaussian RBF} ($\sigma=1.0$). Table~\ref{tab:kernel-ablation} shows that no single kernel is universally optimal; the best choice depends on the dataset and its underlying distribution. Table~1 reports the best performance across these kernels for each task. For a fully fair kernel-by-kernel comparison against the baselines, we extend this to five kernels (Linear, RBF, Laplacian, Mat\'{e}rn-3/2, Rational Quadratic) on all four frontier LLMs in App.~\ref{app:kernel_frontier} (Table~\ref{tab:kernel_frontier}). Two patterns emerge: (i) KNN is largely insensitive to the kernel (since it only enters pairwise similarity), and CEIL shows moderate kernel sensitivity, whereas (ii) \textsc{Kite}'s best kernel achieves the highest accuracy among \emph{all} method--kernel combinations on every frontier model, and even \textsc{Kite}'s worst-case kernel matches or exceeds the best CEIL configuration. This indicates that \textsc{Kite}'s information-theoretic objective is robust to kernel misspecification while still benefiting from a good kernel choice.

\begin{table}[t]  
\centering
\small
\setlength{\tabcolsep}{2pt}

\begin{tabular}{@{}lccc@{}}
\toprule
\textbf{Dataset} & \textbf{Linear} & \textbf{Polynomial} & \textbf{Gaussian RBF} \\
\midrule
SST-5     & 47.95 & 48.38 & \textbf{49.59} \\
CMSQA     & 69.86 & 71.06 & \textbf{71.25} \\
MRPC      & \textbf{75.27} & 71.22 & 67.15 \\
QNLI      & 70.46 & 70.38 & \textbf{71.05} \\
HellaSwag & 67.18 & \textbf{71.02} & 70.67 \\
\bottomrule
\end{tabular}
\vskip -2mm
\caption{\textbf{Ablation study on the kernel function.} We report accuracy for \textsc{Kite} using different kernels on Qwen-1.5B.}
\label{tab:kernel-ablation}
\vspace{-4mm}

\end{table}

\begin{table}[t]
\centering
\small
\setlength{\tabcolsep}{3pt}
\begin{tabular}{@{}lccccc@{}}
\toprule
\textbf{Method} & \textbf{BERT} & \textbf{RoBERTa} & \textbf{E5} & \textbf{DistilBERT} & \textbf{MiniLM} \\
\midrule
CEIL          & 47.88 & 41.41 & 45.68 & 48.77 & 48.13 \\
Dense         & 46.64 & 36.33 & 45.14 & 49.04 & 48.68 \\
\textsc{Kite} & \textbf{49.59} & \textbf{48.31} & \textbf{48.04} & \textbf{49.29} & \textbf{48.72} \\
\bottomrule
\end{tabular}
\vskip -2mm
\caption{\textbf{Robustness to embedding encoders.} Accuracy (\%) on SST-5 with Qwen-2.5-1.5B using different embedding encoders for retrieval. The headers denote \texttt{bert-base-uncased}, \texttt{roberta-base}, \texttt{e5-base-v2}, \texttt{distilbert-base-uncased}, and \texttt{all-MiniLM-L6-v2}, respectively.}
\label{tab:embedding-ablation}
\vspace{-4mm}
\end{table}

\vspace{0.3cm}
\noindent\textbf{Robustness to embedding encoders.} A natural question is whether \textsc{Kite}'s gains are sensitive to the embedding encoder. To investigate, we evaluate \textsc{Kite}, Dense, and CEIL on SST-5 using Qwen-2.5-1.5B with five embedding models: BERT~\cite{devlin2019bertpretrainingdeepbidirectional}, RoBERTa~\citep{liu2019roberta}, E5~\citep{wang2022text}, DistilBERT~\citep{sanh2019distilbert}, and MiniLM~\citep{wang2020minilm}. As shown in Table~\ref{tab:embedding-ablation}, \textsc{Kite} achieves the highest or near-highest accuracy across all encoders. While Dense and CEIL exhibit significant variance (e.g., CEIL drops from $48.77\%$ with DistilBERT to $41.41\%$ with RoBERTa), \textsc{Kite} remains consistently strong, with a much narrower performance range. This shows that our information-theoretic selection criterion is robust to the embedding backbone, underscoring \textsc{Kite}'s practical deployability. We further confirm this with the NV-Embed-v2~\citep{lee2024nvembed} encoder across all four frontier LLMs and seven classification datasets (App.~\ref{app:nvembed}, Table~\ref{tab:nvembed_summary}): \textsc{Kite} attains the highest average on all models, with gains of $+1.0$ to $+2.1$ points over the best KNN retriever.

\vspace{0.3cm}
\noindent\textbf{Per-dataset analysis.} Table~\ref{tab:frontier_perdataset} provides per-dataset results for two representative models.
On GPT-4.1, \textsc{Kite} achieves substantial gains on SWAG ($+3.2$ over both KNN \& CEIL), MRPC ($+2.4$ over KNN), and QNLI ($+0.8$ over KNN). On Claude-Haiku-4.5, the largest gains appear on SWAG ($+3.0$ over KNN), QNLI ($+1.6$), and MRPC ($+1.5$). On near-saturated benchmarks (IMDB, DBPedia), performance remains tied or within noise, as expected. Across all four models and seven datasets, \textsc{Kite} matches or exceeds the best baseline in the majority of cases, with remaining gaps typically within $0.5\%$ (see Appendix~\ref{app:perdataset_frontier} for full per-model breakdowns).

\noindent\textbf{Extension to QA and reasoning tasks.} We further evaluate \textsc{Kite} on four QA and reasoning benchmarks: NQ~\citep{kwiatkowski-etal-2019-natural}, TriviaQA~\citep{joshi2017triviaqa}, HotpotQA~\citep{yang2018hotpotqa}, and GSM8K~\citep{cobbe2021training}, using four frontier LLMs with $r{=}5$. NQ, TriviaQA, and HotpotQA are scored with Soft Exact Match (normalised substring overlap); GSM8K uses chain-of-thought prompting  and numeric match on the final answer. As shown in Table~\ref{tab:qa_crossmodel} (Appendix~\ref{app:qa_results}), \textsc{Kite} achieves the highest average accuracy on all four models, with gains of $+0.2$ to $+1.8$ over KNN.

\begin{table}[h]
\centering
\setlength{\tabcolsep}{3.5pt}
\resizebox{\linewidth}{!}{%
\begin{tabular}{l ccc cc | ccc cc}
\toprule
& \multicolumn{5}{c}{\textbf{GPT-4.1}} & \multicolumn{5}{c}{\textbf{Claude-Haiku-4.5}} \\
\cmidrule(lr){2-6} \cmidrule(lr){7-11}
\textbf{Dataset} & \textbf{KNN} & \textbf{CEIL} & \textbf{\textsc{Kite}} & $\Delta_{\text{K}}$ & $\Delta_{\text{D}}$
                 & \textbf{KNN} & \textbf{CEIL} & \textbf{\textsc{Kite}} & $\Delta_{\text{K}}$ & $\Delta_{\text{D}}$ \\
\midrule
SST-5     & 53.6 & 52.8 & \textbf{54.2} & +0.6 & +1.4  & 52.8 & 53.2 & \textbf{54.0} & +1.2 & +0.8 \\
MRPC      & 68.9 & 70.1 & \textbf{71.3} & +2.4 & +1.2  & 71.8 & 72.8 & \textbf{73.3} & +1.5 & +0.5 \\
QNLI      & 83.8 & 83.0 & \textbf{84.6} & +0.8 & +1.6  & 86.6 & 87.0 & \textbf{88.2} & +1.6 & +1.2 \\
SWAG      & 71.8 & 71.8 & \textbf{75.0} & +3.2 & +3.2  & 75.8 & 76.6 & \textbf{78.8} & +3.0 & +2.2 \\
IMDB      & \textbf{99.4} & 99.2 & \textbf{99.4} & 0.0 & +0.2  & 97.6 & \textbf{97.4} & 97.4 & +0.2 & 0.0 \\
DBPedia   & \textbf{99.0} & \textbf{99.4} & 99.0 & 0.0 & + 0.4  & 99.0 & \textbf{99.6} & \textbf{99.6} & +0.6 & 0.0 \\
BoolQ     & 91.6 & 91.8 & \textbf{92.2} & +0.6 & +0.4  & 92.8 & 93.2 & \textbf{93.0} & +0.2 & +0.2 \\
\midrule
\textbf{Avg} & 81.2 & 81.2 & \textbf{82.2} & +1.0 & +1.0  & 82.3 & 82.8 & \textbf{83.5} & +1.2 & +0.7 \\
\bottomrule
\end{tabular}}
\vskip -2mm
\caption{Accuracy (\%) on core benchmarks for GPT-4.1 and Claude-Haiku-4.5 ($r{=}5$, OpenAI large text embeddings). Best result in \textbf{bold}. $\Delta$ columns show \textsc{Kite}'s gain over each baseline.}
\label{tab:frontier_perdataset}
\vspace{-4mm}
 \end{table}

\begin{table}[h]
\centering
\resizebox{\columnwidth}{!}{%
\begin{tabular}{lcccc c}
\toprule
\textbf{Model} & \textbf{BM25} & \textbf{KNN-best} & \textbf{DPP-best} & \textbf{\textsc{Kite}-best} & $\boldsymbol{\Delta}$\textbf{\textsc{Kite}--KNN} \\
\midrule
GPT-4o-mini        & 64.1 & 65.1 & 65.1 & \textbf{65.3} & \textcolor{green!60!black}{+0.2} \\
GPT-4.1            & 71.2 & 71.5 & 71.7 & \textbf{72.5} & \textcolor{green!60!black}{+1.0} \\
Claude-Haiku-4.5   & 59.1 & 60.3 & 61.0 & \textbf{62.1} & \textcolor{green!60!black}{+1.8} \\
GPT-5.2            & 68.2 & 69.8 & 70.1 & \textbf{71.0} & \textcolor{green!60!black}{+1.2} \\
\bottomrule
\end{tabular}}
\vskip -2mm
\caption{Average accuracy (\%) on four QA/reasoning datasets with frontier LLMs ($r{=}5$, OpenAI large text embeddings). \textsc{Kite}-best = max over all kernels and $\lambda$ values.}
\label{tab:qa_crossmodel}
\vspace{-4mm}
\end{table}

\vspace{0.3cm}
\noindent\textbf{Effect of exemplar budget ($r$).} Note that redundancy is more likely with larger $r$ (more in-context examples). We repeat the frontier-LLM evaluation at $r{=}10$ (Appendix~\ref{app:k10}, Tables~\ref{tab:k10_summary}--\ref{tab:k10_perdataset}). \textsc{Kite}'s advantage over KNN \emph{grows} with budget: from $+1.0$ at $r{=}5$ to $+1.9$ at $r{=}10$ on GPT-4.1, and from $+1.2$ to $+2.1$ on Claude-Haiku-4.5 (App.~\ref{app:k5_vs_k10}, Table~\ref{tab:k5vsk10}). Per-dataset, the largest $r{=}10$ gains over KNN appear on SWAG ($+3.8$ on GPT-4.1, $+3.0$ on Claude-Haiku-4.5), QNLI ($+2.4$ and $+4.2$), and SST-5 ($+4.4$ and $+3.2$). At the other end of the budget spectrum, App.~\ref{app:lambda_ice_ablation} (Figure~\ref{fig:ice_ablation}) shows that on SST-5 the gap between \textsc{Kite} and BM25/Dense/CEIL is largest in the low-shot regime ($r{=}2$--$10$), where careful selection matters most. These indicate that \textsc{Kite} is effective across the full budget range.

\vspace{0.5mm}
\section{Conclusion}

We introduced \textsc{Kite}, a query-specific greedy exemplar selection method for ICL. It selects examples that balance relevance and diversity, yielding consistent improvements over strong retrieval baselines across multiple datasets and models.

\section*{Limitations}

In Section~\ref{sec:exp}, we demonstrated the effectiveness of our proposed framework across multiple models and benchmarks. However, our theoretical analysis is currently grounded in classification tasks. Extending the theoretical analysis of \textsc{Kite} to open-ended generation tasks such as summarization or dialogue requires defining an appropriate theoretical surrogate objective, which we leave as future work.

A second consideration is computational scaling. \textsc{Kite}'s greedy loop is $O(r \cdot n)$ per query, which is comparable to DPP-based methods and incurs negligible overhead at the bank sizes used in our experiments (up to $n \approx 8{,}500$; Appendix~\ref{app:runtime}). For substantially larger candidate banks (e.g., $n \gg 10^4$, as in MNLI's $\approx 400{,}000$ examples), \textsc{Kite} is naturally combined with an approximate-nearest-neighbor pre-filter to obtain a top-$N$ candidate pool ($N \ll n$) before the greedy selection step. The approximation guarantee of Theorem~\ref{thm:greedy} then applies with respect to this pruned ground set; characterizing the trade-off between $N$ and the resulting approximation quality is an interesting direction for future work.

\clearpage
\newpage
\vspace{-5mm}
\bibliography{main}

\clearpage
\newpage

\appendix

\section{Extended Ablation Study}
\label{app:lambda_ice_ablation}

\begin{figure}[!h]
    \centering
    \includegraphics[width=0.9 \linewidth]{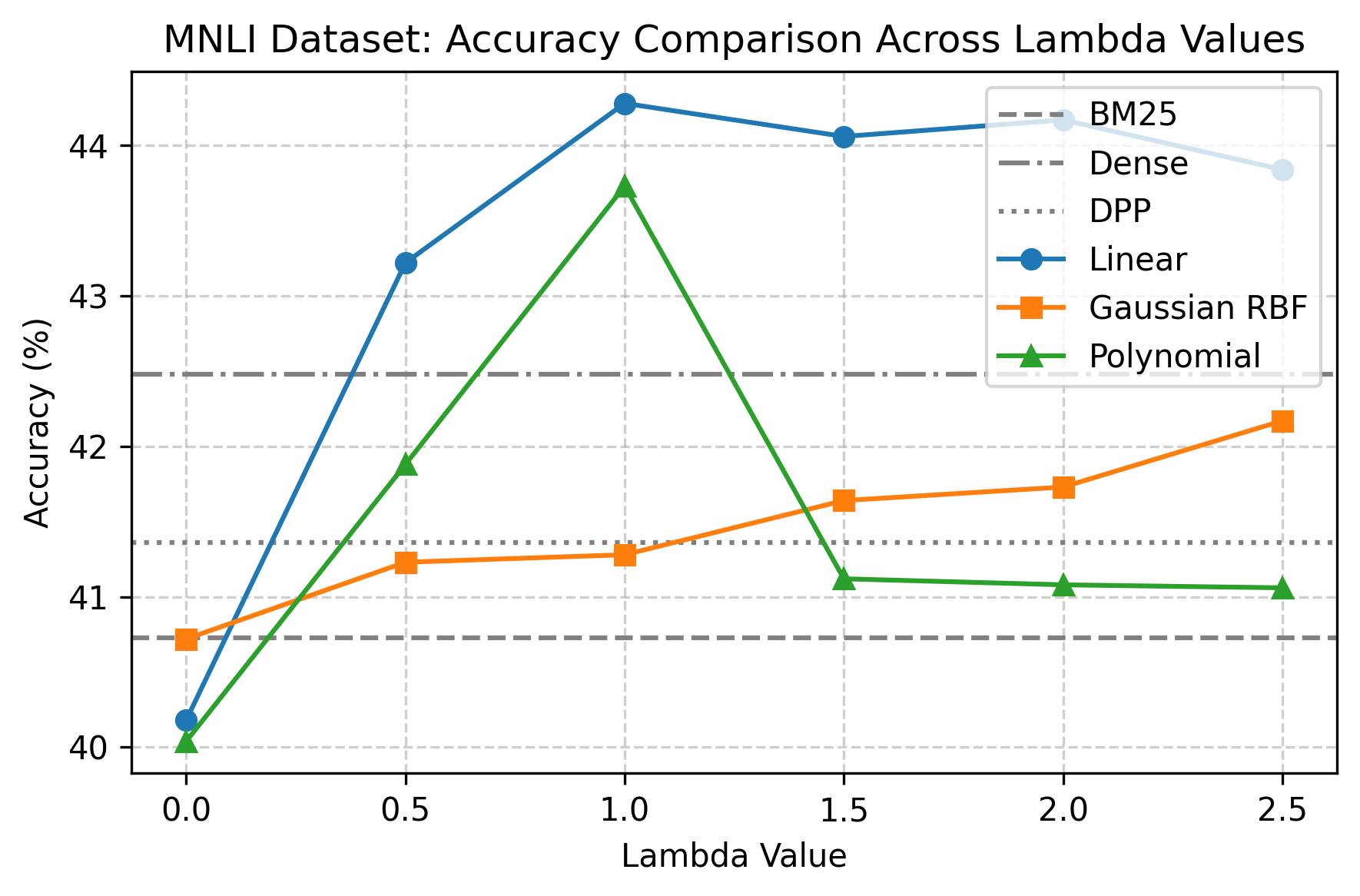}

    \caption{Ablation study on the MNLI dataset illustrating the effect of varying $\lambda$ values. We report accuracy for KITE using different kernel-based selection strategies on the Qwen-1.5B model. Since MNLI has a large example bank (over 400k examples), a purely relevance-based selection ($\lambda = 0$) underperforms. Incorporating diversity through higher $\lambda$ values (e.g., $\lambda = 1$) leads to notable improvements in accuracy by promoting a more diverse selection of in-context examples.}
        \label{fig:mnli_ablation}
\end{figure}

\paragraph{Ablation study on $\lambda$.} In this section, we present an ablation study over $\lambda$ on the MNLI dataset~\cite{williams2017broad}. As shown in Figure~\ref{fig:mnli_ablation}, we evaluate the impact of the trade-off parameter $\lambda$, which balances relevance and diversity in our selection process. For the MNLI dataset, which has a very large and varied example bank (over 400k examples), relying solely on relevance-based selection ($\lambda$=0) results in suboptimal performance. This is because many examples can be semantically similar, leading to a redundant in-context set. By increasing $\lambda$, we introduce diversity into the selection, which proves crucial for this task. The results indicate that a balanced approach, with $\lambda$ set to approximately 1, achieves the highest accuracy. This demonstrates that for complex, large-scale datasets, a combination of both relevance and diversity is essential for selecting the most effective in-context examples.

\begin{figure}[!h]
    \centering
    \includegraphics[width=0.9 \linewidth]{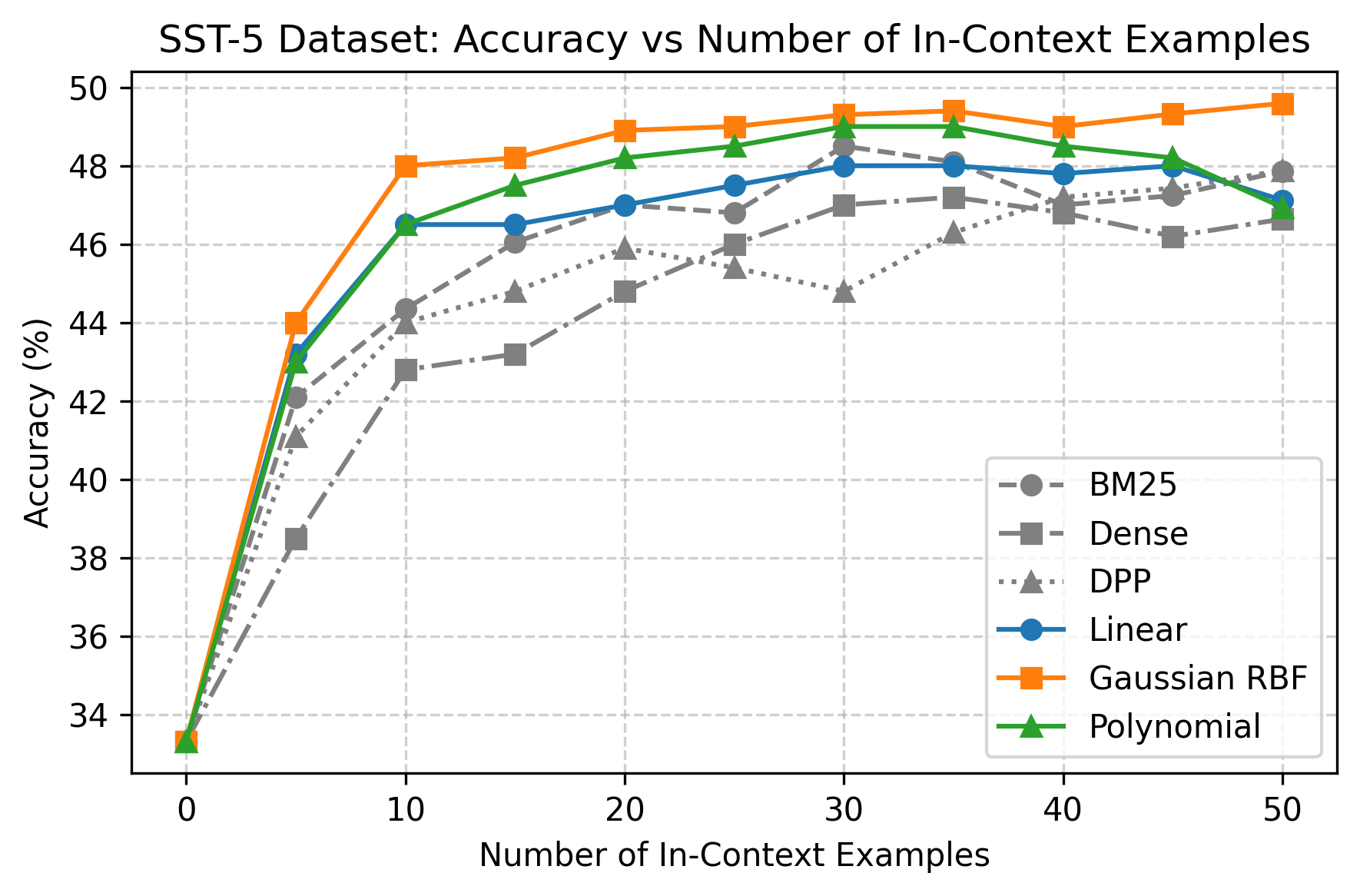}
    \caption{Accuracy comparison of the Qwen-1.5B model on the SST-5 dataset across varying numbers of in-context examples (ICE). The KITE algorithm, using kernel-based selection strategies, significantly outperforms baseline methods (BM25, Dense, DPP) in low-ICE settings.}
    \label{fig:ice_ablation}
\end{figure}

\paragraph{Ablation study on number of ICE examples.} We also experimented with the number of in-context examples to understand how it affects model performance. In Figure~\ref{fig:ice_ablation}, we report the accuracy of the Qwen-1.5B model on the SST-5 dataset as a function of the number of provided examples. As expected, performance for all methods generally improves as more examples are added to the context. However, our proposed \textsc{KITE} algorithm consistently and significantly outperforms all baseline methods, including BM25, Dense retrieval, and DPP. The advantage of KITE is particularly pronounced in low-resource settings (e.g., with 2 or 4 ICE), where it establishes a substantial performance gap. This highlights \textsc{KITE}'s superior ability to select high-quality, informative examples, making it highly effective even when the number of in-context examples is limited.

\section{Empirical Validation of Submodularity}
\label{app:submodularity}

\newcolumntype{?}{!{\vrule width 1pt}}
\begin{table*}[!t]
\centering
\setlength{\tabcolsep}{2pt}
\resizebox{0.8\textwidth}{!}{
\begin{tabular}{lccccccc ? cccccc}
\toprule
& \multicolumn{7}{c?}{\textbf{Vary \(r\) (fixed \(\beta=1\))}} & \multicolumn{6}{c}{\textbf{Vary \(\beta\) (fixed \(r=20\))}} \\
\cmidrule(lr){2-8}\cmidrule(lr){9-14}
\textbf{Dataset} & 5 & 10 & 15 & 20 & 25 & 30 & 35 & 1 & 3 & 5 & 7 & 9 & 11 \\
\midrule
SST\textendash5   & 0.748 & 0.685 & 0.705 & 0.689 & 0.703 & 0.716 & 0.696 & 0.689 & 0.744 & 0.719 & 0.758 & 0.792 & 0.827 \\
CMSQA             & 0.852 & 0.811 & 0.816 & 0.841 & 0.830 & 0.797 & 0.836 & 0.841 & 0.804 & 0.884 & 0.843 & 0.907 & 0.886 \\
MRPC             & 0.876 & 0.840 & 0.793 & 0.816 & 0.850 & 0.887 & 0.848 & 0.816 & 0.765 & 0.803 & 0.824 & 0.824 & 0.898 \\
QNLI             & 0.844 & 0.801 & 0.900 & 0.893 & 0.875 & 0.837 & 0.877 & 0.893 & 0.829 & 0.823 & 0.919 & 0.912 & 0.882 \\
HellaSwag        & 0.395 & 0.601 & 0.525 & 0.596 & 0.839 & 0.908 & 0.806 & 0.596 & 0.803 & 0.767 & 0.806 & 0.935 & 0.852 \\

\bottomrule
\end{tabular}
}
\caption{\textbf{Analysis on Submodularity ratio \(\gamma_{\min}\)}. \textbf{Left:} We vary \(r\) while fixing \(\beta = 1\); \textbf{Right:} We vary \(\beta\) while fixing \(r = 20\).}
\label{tab:gamma}
\end{table*}

We empirically validate the submodularity of our set function, $f_z(\mathcal{S})$ by estimating its submodularity ratio (Def.~\ref{def:submod_ratio}) on real-world text embeddings. For this analysis, we used the first 500 examples from each dataset to generate embeddings. We created two distinct types: demonstration embeddings, which concatenate an input with its gold label, and {query embeddings}, which use the input alone. 

To estimate the ratio, we run a Monte Carlo simulation across a grid of hyperparameters for subset size $r$ and regularization $\beta$. In each trial, we sample a disjoint triplet $(\mathcal{S}, \mathcal{L}, \mathbf{z})$, where $\mathcal{S}$ is a random set of demonstration embeddings, $\mathcal{L}$ is a diverse set of up to $r$ demonstration embeddings selected via farthest-point sampling, and $\mathbf{z}$ is a query embedding. We then compute the ratio and record the minimum value observed for each configuration ($\gamma_{\min}$). As shown in Table~\ref{tab:gamma}, $\gamma_{\min}$ is typically in the $0.7$--$0.9$ range across most datasets and settings, and approaches $1$ as $r$ or $\beta$ grow (right block of Table~\ref{tab:gamma}). The lowest values appear on HellaSwag at small $r$ (down to $0.395$ at $r{=}5$, $\beta{=}1$), where the candidate bank's higher embedding coherence reduces the ratio. Even in that regime, greedy selection remains empirically strong (Table~\ref{tab:classification-results}); the theoretical guarantee in Theorem~\ref{thm:greedy} degrades gracefully with $\gamma_r$, so a $\gamma$ of $0.4$ still yields a $\approx 0.33$-factor approximation rather than no guarantee. Overall, the data support the use of a greedy algorithm. This finding justifies our use of a greedy algorithm, as it is expected to yield near-optimal results.

\section{Theoretical Proofs and Results}

In this section, we present the missing proofs.

\subsection{Kernel Trick}
The following well-known result on linear operators helps us apply the kernel trick to compute the greedy selection rule of \textsc{Kite}, see~\eqref{eq:kernel-trick}.
\begin{lemma}
   \label{lem:dim-change}
   Let $\fl{A}$ be a linear operator. Then, for any $\beta > 0$, the following holds:
   \begin{align*}
     (\fl{A}^{\top} \fl{A} + \beta \fl{I})^{-1}\fl{A}^{\top}&=\fl{A}^{\top}(\fl{A}\fl{A}^{\top}+\beta \fl{I})^{-1},\\
   \fl{I} - \fl{A}^{\top}(\fl{A}\fl{A}^{\top}+\beta \fl{I})^{-1}\fl{A} &= \beta (\fl{A}^{\top} \fl{A} + \beta \fl{I})^{-1}.  
   \end{align*}
\end{lemma}
\noindent
Using these identities, we get
\begin{align*}
&\phi(\fl{z})^{\top}\mathbf{V}_{\mathcal S}^{-1}\phi(\fl{z}) = \phi(\fl{z})^{\top}(\Phi_{\mathcal S}^{\top}\Phi_{\mathcal S}+\beta \mathbf{I})^{-1}\phi(\fl{z}) \nonumber\\
&= \frac{1}{\beta}\Big(
\phi(\fl{z})^{\top}\phi(\fl{z}) \nonumber\\
&- \phi(\fl{z})^{\top}\Phi_{\mathcal S}^{\top}
(\Phi_{\mathcal S}\Phi_{\mathcal S}^{\top}+\beta \mathbf{I}_{|\ca{S}|})^{-1}
\Phi_{\mathcal S}\phi(\fl{z})
\Big) \nonumber\\
&= \frac{1}{\beta}\Big(
k(\fl{z},\fl{z})
- \mathbf{k}_{\mathcal S}(\fl{z})^{\top}
(\mathbf{K}_{\mathcal S}+\beta \mathbf{I}_{|\ca{S}|})^{-1}
\mathbf{k}_{\mathcal S}(\fl{z})
\Big)\nonumber\\
&= \frac{1}{\beta}k_{\ca{S}}(\fl{z},\fl{z})~.
\end{align*}

\subsection{Proof of Lemma \ref{lem:submod}}

For a set \( \mathcal{S} \subseteq \mathcal{X} \) and an element \( \mathbf{x}_i \notin \mathcal{S} \), define the marginal gain of adding \( \mathbf{x}_i \) to \( \mathcal{S} \) in the set function $f_{\mathbf{z}}(\mathcal{S})$ as (the second step is obtained via an application of the Sherman Morrison formula)
\begin{align*}
\Delta_i 
&:= f_{\mathbf{z}}(\mathcal{S} \cup \{\mathbf{x}_i\}) - f_{\mathbf{z}}(\mathcal{S}) \\
&= 
\frac{\big(\phi(\mathbf{z})^\top \mathbf{V}_{\mathcal{S}}^{-1} \phi(\mathbf{x}_i)\big)^2}
{1 + \phi(\mathbf{x}_i)^\top \mathbf{V}_{\mathcal{S}}^{-1} \phi(\mathbf{x}_i)},
\end{align*}
where $f_{\mathbf{z}}(\mathcal{S}) 
= -\phi(\mathbf{z})^\top \mathbf{V}_{\mathcal{S}}^{-1} \phi(\mathbf{z})$
and $\mathbf{V}_{\mathcal{S}} 
= \sum_{i \in \mathcal{S}} \phi(\mathbf{x}_i)\phi(\mathbf{x}_i)^\top + \beta \mathbf{I}$.
Now consider a set $\mathcal{L} = \{\mathbf{x}_1,\dots,\mathbf{x}_r\}
\subseteq \mathcal{X} \setminus \mathcal{S}$. We compare
\[
\sum_{\mathbf{x}_i \in \mathcal{L}} \Delta_i,
\qquad \text{vs.} \qquad
f_{\mathbf{z}}(\mathcal{S} \cup \mathcal{L}) - f_{\mathbf{z}}(\mathcal{S}).
\]
The ratio of these two quantities defines the \emph{submodularity ratio} \( \gamma \), which quantifies how closely the function $f$ adheres
to submodularity. Let
\[
\Phi_{\mathcal{L}} =
\begin{bmatrix}
\phi(\mathbf{x}_1)^\top \\
\phi(\mathbf{x}_2)^\top \\
\vdots \\
\phi(\mathbf{x}_r)^\top
\end{bmatrix}
\]
be the matrix obtained by stacking the feature vectors corresponding to the elements of \( \mathcal{L} \).
Then the updated covariance matrix becomes
\[
\mathbf{V}_{\mathcal{S} \cup \mathcal{L}}
=
\mathbf{V}_{\mathcal{S}} + \Phi_{\mathcal{L}}^\top \Phi_{\mathcal{L}}.
\]
To compute \( \mathbf{V}_{\mathcal{S} \cup \mathcal{L}}^{-1} \), we apply the Woodbury matrix identity and obtain
\begin{align*}
\mathbf{V}_{\mathcal{S} \cup \mathcal{L}}^{-1}
&=
\mathbf{V}_{\mathcal{S}}^{-1}
- \\
&\mathbf{V}_{\mathcal{S}}^{-1}
\Phi_{\mathcal{L}}^\top
\left(
\mathbf{I} + \Phi_{\mathcal{L}}\mathbf{V}_{\mathcal{S}}^{-1}\Phi_{\mathcal{L}}^\top
\right)^{-1}
\Phi_{\mathcal{L}}
\mathbf{V}_{\mathcal{S}}^{-1}.
\end{align*}
Substituting this into the objective gives
\begin{align*}
&f_{\mathbf{z}}(\mathcal{S} \cup \mathcal{L}) =
-\phi(\mathbf{z})^\top
\mathbf{V}_{\mathcal{S} \cup \mathcal{L}}^{-1}
\phi(\mathbf{z}) \\
&=
-\phi(\mathbf{z})^\top
\mathbf{V}_{\mathcal{S}}^{-1}
\phi(\mathbf{z}) \\
&+
\phi(\mathbf{z})^\top
\mathbf{V}_{\mathcal{S}}^{-1}
\Phi_{\mathcal{L}}^\top
\left(
\mathbf{I} + \Phi_{\mathcal{L}}\mathbf{V}_{\mathcal{S}}^{-1}\Phi_{\mathcal{L}}^\top
\right)^{-1}
\Phi_{\mathcal{L}}
\mathbf{V}_{\mathcal{S}}^{-1}
\phi(\mathbf{z}).
\end{align*}
Hence the total gain from adding \( \mathcal{L} \) is
\begin{align*}
&f_{\mathbf{z}}(\mathcal{S} \cup \mathcal{L}) - f_{\mathbf{z}}(\mathcal{S}) \\
&=
\phi(\mathbf{z})^\top
\mathbf{V}_{\mathcal{S}}^{-1}
\Phi_{\mathcal{L}}^\top
\left(
\mathbf{I} + \Phi_{\mathcal{L}}\mathbf{V}_{\mathcal{S}}^{-1}\Phi_{\mathcal{L}}^\top
\right)^{-1}
\Phi_{\mathcal{L}}
\mathbf{V}_{\mathcal{S}}^{-1}
\phi(\mathbf{z}).
\end{align*}
Let $\mathbf{w} = \mathbf{V}_{\mathcal{S}}^{-1}\phi(\mathbf{z})$. Then, we have
\begin{align*}
&f_{\mathbf{z}}(\mathcal{S} \cup \mathcal{L}) - f_{\mathbf{z}}(\mathcal{S}) \\
&=
\mathbf{w}^\top
\Phi_{\mathcal{L}}^\top
\left(
\mathbf{I} + \Phi_{\mathcal{L}}\mathbf{V}_{\mathcal{S}}^{-1}\Phi_{\mathcal{L}}^\top
\right)^{-1}
\Phi_{\mathcal{L}}
\mathbf{w}~,
\end{align*}
which can be interpreted as a bilinear form involving the projection of the vector $\mathbf{w}$ onto the subspace spanned by the rows of $\Phi_{\mathcal{L}}$, that is, the set of vectors $\{\phi(\mathbf{x}_j)\}_{j \in \mathcal{L}}$. The matrix \(\big(\mathbf{I}+\Phi_{\mathcal L}\mathbf{V}_{\mathcal S}^{-1}\Phi_{\mathcal L}^\top\big)^{-1}\) acts as a correction factor that accounts for interactions between the new feature directions \(\{\phi(\mathbf{x}_j)\}_{j\in\mathcal L}\) and the current inverse covariance \(\mathbf{V}_{\mathcal S}^{-1}\). Intuitively, the total gain from adding a group of elements depends not only on the alignment of each \(\phi(\mathbf{x}_j)\) with \(\mathbf{w}\) but also on their mutual (in)dependence in the \(\mathbf{V}_{\mathcal S}^{-1}\)-induced geometry.

Let us define the matrix $\mathbf{G}
=
\Phi_{\mathcal{L}}
\mathbf{V}_{\mathcal{S}}^{-1}
\Phi_{\mathcal{L}}^\top
=
\mathbf{D} + \mathbf{N}$,
where $\mathbf{D}
=
\operatorname{diag}\!\big(
\phi(\mathbf{x}_j)^\top
\mathbf{V}_{\mathcal{S}}^{-1}
\phi(\mathbf{x}_j)
\big)$
and \( \mathbf{N} \) contains the off-diagonal terms $\big(
\phi(\mathbf{x}_j)^\top
\mathbf{V}_{\mathcal{S}}^{-1}
\phi(\mathbf{x}_k)
\big)_{j \neq k}$.
We ideally want these terms to be low, i.e., low mutual coherence among the rows of \( \Phi_\ca{L} \) in the \( \fl{V}_\ca{S}^{-1} \)-inner product. Now see that
\begin{align*}
&(\fl{D} + \fl{N})^{-1} = \left( \fl{D} \left( \fl{I} + \fl{D}^{-1}\fl{N} \right) \right)^{-1} \\
&= \left( \fl{I} + \fl{D}^{-1}\fl{N} \right)^{-1} \fl{D}^{-1}.
\end{align*}
If \( \|\mathbf{D}^{-1}\mathbf{N}\| \ll 1 \), then we can expand the inverse using a Neumann series \footnote{This condition holds when vectors \( \phi(\fl{x}_i) \) and \( \phi(\fl{x}_j) \) are nearly orthogonal under the \( \fl{V}_\ca{S}^{-1} \)-induced geometry, which ensures that the off-diagonal elements of \( \fl{G} \) are small relative to the diagonal ones.}
\[
\left( \fl{I} + \fl{D}^{-1}\fl{N} \right)^{-1} = \sum_{k=0}^{\infty} (-1)^k (\fl{D}^{-1}\fl{N})^k.
\]
Then we obtain
\[
(\fl{D} + \fl{N})^{-1} = \left( \sum_{k=0}^{\infty} (-1)^k (\fl{D}^{-1}\fl{N})^k \right) \fl{D}^{-1}.
\]  
Keeping only the first two terms in the series, we get $(\fl{D} + \fl{N})^{-1} \approx \fl{D}^{-1} - \fl{D}^{-1}\fl{N} \fl{D}^{-1}$.
This approximation is accurate when the norm of \( \fl{N} \) is small relative to \( \fl{D} \), i.e., \( \| \fl{D}^{-1}\fl{N} \|=\| \fl{D}^{-1/2} \fl{N} \fl{D}^{-1/2}\| \ll 1 \) as $\fl{N}$ is a symmetric matrix.
Therefore, we get
\begin{align*}
&(\fl{I} + \fl{G})^{-1} \approx \operatorname{diag}\left( \frac{1}{1 + \phi(\fl{x}_j)^\top \fl{V}_\ca{S}^{-1} \phi(\fl{x}_j)} \right) \\ 
&+ \fl{D}^{-1}\fl{N} \fl{D}^{-1}.
\end{align*}
Now, note that
\begin{align*}
&\fl{w}^\top \fl{\Phi}_\ca{L}^\top\operatorname{diag}\left( \frac{1}{1 + \phi(\fl{x}_j)^\top \fl{V}_\ca{S}^{-1} \phi(\fl{x}_j)} \right) \Phi_\ca{L} \fl{w}\\ &= \sum_{i=1}^r \frac{(\phi(\fl{z})^\top \fl{V}_\ca{S}^{-1} \phi(\fl{x}_i))^2}{1 + \phi(\fl{x}_i)^\top \fl{V}_\ca{S}^{-1} \phi(\fl{x}_i)} = \sum_{i=1}^r \Delta_i~.
\end{align*}
Similarly, for the diagonal matrix
$\fl{D} = \operatorname{diag}(d_1, d_2, \dots, d_n)$ and the off-diagonal matrix
 \( \fl{N} = [n_{ij}] \in \mathbb{R}^{n \times n} \), we have
 \[
\fl{w}^\top \fl{D}^{-1}\fl{N} \fl{D}^{-1} \fl{w} = \sum_{i \neq j} \frac{w_i w_j}{d_i d_j} n_{ij}.
\]
This gives us
\begin{align*}
  & \fl{w}^\top \Phi^\top_\ca{L} \fl{D}^{-1}\fl{N} \fl{D}^{-1} \Phi_\ca{L} \fl{w}\\ &= \sum_{ i \neq j} \Big(\frac{\phi(\fl{z})^\top \fl{V}_\ca{S}^{-1} \phi(\fl{x}_i)}{1 + \phi(\fl{x}_i)^\top \fl{V}_\ca{S}^{-1} \phi(\fl{x}_i)}\Big) \\ & \times \Big(\frac{\phi(\fl{z})^\top \fl{V}_\ca{S}^{-1} \phi(\fl{x}_j)}{1 + \phi(\fl{x}_j)^\top \fl{V}_\ca{S}^{-1} \phi(\fl{x}_j)} \Big) 
  \times ( \phi(\fl{x}_i)^\top \fl{V}_\ca{S}^{-1} \phi(\fl{x}_j)) \\
  & = \sum_{ i \neq j} \sqrt{\Delta_i \Delta_j} \mu_{i,j}~,
\end{align*}
where $\mu_{ij}$ is given by
\[
\frac{
\phi(\mathbf{x}_i)^\top \mathbf{V}_{\mathcal{S}}^{-1}\phi(\mathbf{x}_j)
}{
\sqrt{1+\phi(\mathbf{x}_i)^\top \mathbf{V}_{\mathcal{S}}^{-1}\phi(\mathbf{x}_i)}
\sqrt{1+\phi(\mathbf{x}_j)^\top \mathbf{V}_{\mathcal{S}}^{-1}\phi(\mathbf{x}_j)}
}.
\]
Therefore, we obtain
\[
f_\fl{z}(\ca{S} \cup \ca{L}) - f_\fl{z}(\ca{S}) = \sum_{i=1}^k \Delta_i -  \sum_{ i \neq j} \sqrt{\Delta_i \Delta_j} \mu_{i,j}~.
\]
This implies
\[
\gamma_r(f,\mathbf{z},\mathcal{L},\mathcal{S})
=
\frac{\sum_{i=1}^r \Delta_i}
{\sum_{i=1}^r \Delta_i - \sum_{i\ne j}\sqrt{\Delta_i\Delta_j}\mu_{ij}}.
\]
Let us denote $a_i := \sqrt{\Delta_i}$. Then we have
\[
\gamma_r(f,\fl{z},\ca{L},\ca{S}) = \frac{\sum\limits_{i=1}^r a_i^2}{\sum\limits_{i=1}^{r} a_i^2 - \sum\limits_{i \ne j} a_i a_j \mu_{i,j}}.
\]
Using the bound $|\mu_{i,j} | \le \mu$, we get
\begin{align*}
    &| \sum_{i \ne j} a_i a_j \mu_{i,j} | \le \mu \sum_{i \ne j}  | a_i a_j |\\
    &\implies -\sum_{i \ne j} a_i a_j \mu_{i,j} \le \mu \sum_{i \ne j}  | a_i a_j |~.
\end{align*}
 To bound $\sum_{i \ne j} |a_i a_j|$, note that
\begin{align*}
&\left( \sum_{i=1}^r |a_i| \right)^2 = \sum_{i=1}^r a_i^2 + \sum_{i \ne j} |a_i a_j|\\
& \implies \sum_{i \ne j} |a_i a_j | = \left( \sum_{i=1}^r |a_i| \right)^2 - \sum_{i=1}^r a_i^2\\
& \implies \sum_{i \ne j} |a_i a_j | \le (r - 1) \sum_{i=1}^r a_i^2~,
\end{align*}
since by the Cauchy-Schwarz inequality
$\left( \sum_{i=1}^r |a_i| \right)^2 \le r \sum_{i=1}^r a_i^2$.
Therefore, we can bound the denominator as
\[
\sum_{i=1}^r a_i^2 - \sum_{i \ne j} a_i a_j \mu_{i,j}
\le \left(1 + (r - 1)\mu\right) \sum_{i=1}^r a_i^2~,
\]
which yields 
\begin{align*}
  \gamma_r(f,\fl{z},\ca{L},\ca{S}) \ge   \frac{1}{1+(r - 1)\mu}~.
\end{align*}
Since the bound holds for any $\fl{z}$, any $\ca{S} \subseteq \ca{X}$ and any $\ca{L} \subseteq \ca{X}$ such that $\cal{L} \cap \ca{S} = \emptyset$, we can conclude the proof by applying kernel trick to rewrite 
\begin{align*}
    \mu_{i,j}= \frac{k_{\ca{S}}(\fl{x}_i,\fl{x}_j)}{\sqrt{\beta+k_{\ca{S}}(\fl{x}_i,\fl{x}_i)\sqrt{\beta+k_{\ca{S}}(\fl{x}_j,\fl{x}_j)}}}~.
\end{align*}

\subsection{Proof of Theorem \ref{thm:greedy}}

From \citet{das2011submodular}, we know that for any $\fl{z} \in \mathbb{R}^d$, if the greedy algorithm return a set \( \ca{S}_{\text{greedy}} \) for the optimization problem in equation \eqref{eq:set-func}, then
$\ca{S}_{\text{greedy}}$ satisfies
\[
f_{\fl{z}}(\ca{S}_{\text{greedy}}) \geq \left(1 - e^{-\gamma} \right) \cdot f(\ca{S}^\ast),
\]
where \( S^\ast \) is an optimal solution of size at most \( r \). Substituting the lower bound on submodularity ratio from Lemma \ref{lem:submod}, i.e., $\gamma \ge \frac{1}{1+(r - 1)\mu}$, we get the desired bound.

\subsection{Proof of Lemma~\ref{lem:submod_div}}

The result holds by first noting that $\fl{K}_{\mathcal{S}}=\Phi_{\mathcal{S}}\Phi_{\mathcal{S}}^\top$, then applying Sylvester’s identity to get $\det (\fl{K}_\mathcal{S}+\beta \fl{I}_r)=\det \fl{V}_{\ca{S}}$, and finally invoking the matrix-determinant lemma, which, for any $\fl{x} \notin \ca{S}$, yields
\begin{align}\label{eq:inter}
  &\log \det(\fl{V}_\ca{S} + \phi(\fl{x}) \phi(\fl{x})^\top) - \log \det(\fl{V}_\ca{S})\nonumber\\& =\log(1 + \phi(\fl{x})^\top \fl{V}_\ca{S}^{-1} \phi(\fl{x}))\nonumber\\&=\log (1+k_\ca{S}(\fl{x},\fl{x})/\beta).  
\end{align}
Monotonicity follows since this increment \( \log(1 + k_\ca{S}(\fl{x},\fl{x})/\beta) \) is non-negative. Submodularity holds because this increment decreases as the set \( S \) grows, due to redundancy in the directions already spanned. Furthermore, \eqref{eq:inter} naturally admits the greedy selection rule.

\section{Dataset Descriptions}
\label{sec:dataset-details}

Table~\ref{tab:prompt-details} lists the prompt templates and an example input--output pair for each dataset, along with the train/validation split sizes used in our experiments. The text below provides a short description of each.
\textbf{SST-2} is a sentiment classification benchmark containing two coarse-grained classes: \emph{positive} and \emph{negative}.

\noindent
\textbf{SST-5} is a sentiment classification benchmark containing five fine-grained classes: \emph{very positive}, \emph{positive}, \emph{neutral}, \emph{negative}, and \emph{very negative}.

\noindent
\textbf{MRPC} is a corpus of sentence pairs automatically extracted from online news sources, with human annotations indicating whether the sentences in each pair are semantically equivalent.

\noindent
\textbf{MNLI} is a crowdsourced collection of premise--hypothesis sentence pairs annotated for textual entailment. The task is to predict whether the premise \emph{entails}, \emph{contradicts}, or is \emph{neutral} with respect to the hypothesis.

\noindent
\textbf{QNLI} is a question--sentence dataset derived from QA pairs, where the task is to determine whether the given context sentence contains the answer to the question.

\noindent
\textbf{CMSQA} is a multiple-choice question-answering dataset requiring diverse forms of commonsense knowledge. Given a question and five candidate answers, the model must select the correct one.

\noindent
\textbf{HellaSwag} is a large-scale benchmark for grounded commonsense reasoning. Each example pairs a context with four candidate endings: one true video caption (from ActivityNet Captions and the Large Scale Movie Description Challenge) and three adversarially generated distractors designed to fool machines.

\noindent
\textbf{SWAG}~\citep{zellers2018swag} is a multiple-choice sentence-completion benchmark for grounded commonsense inference. Each example consists of a partial sentence followed by four candidate continuations; the model must select the most plausible one. \emph{Note:} the open-source experiments (Table~\ref{tab:classification-results}) use HellaSwag~\citep{zellers-etal-2019-hellaswag}; the frontier LLM experiments (Tables~\ref{tab:frontier_summary}--\ref{tab:k10_perdataset_other}) use SWAG, a related but distinct benchmark from the same authors.

\noindent
\textbf{IMDB} is a binary sentiment classification benchmark of movie reviews. Each review is labeled \emph{positive} or \emph{negative}.

\noindent
\textbf{DBPedia} is a 14-class topic-classification benchmark constructed from Wikipedia article abstracts, with classes spanning company, educational institution, artist, athlete, office holder, mean of transportation, building, natural place, village, animal, plant, album, film, and written work.

\noindent
\textbf{BoolQ} is a yes/no question-answering benchmark. Each example consists of a passage and a question, and the task is to predict whether the answer is \emph{yes} or \emph{no} given the passage.

 
 
\section{Extended Results on Frontier LLMs}
\label{app:frontier_full}
 
We provide complete experimental results for the frontier LLM evaluation described in Section~\ref{sec:frontier}. All experiments use $\beta{=}0.02$ and report the best result across five kernel choices (Linear, RBF, Laplacian, Mat\'{e}rn-3/2, Rational Quadratic) unless otherwise stated.
 
 
\subsection{Per-Model Per-Dataset Results ($r{=}5$, OpenAI Embeddings)}
\label{app:perdataset_frontier}
 
Tables~\ref{tab:app_h2h_4omini}--\ref{tab:app_h2h_gpt52} present per-dataset head-to-head comparisons for all four frontier models.
 
\begin{table}[h]
\centering
\small

\begin{tabular}{lccccc}
\toprule
\textbf{Dataset} & \textbf{\textsc{Kite}} & \textbf{KNN} & \textbf{CEIL} & $\boldsymbol{\Delta}_{\text{KNN}}$ & $\boldsymbol{\Delta}_{\text{CEIL}}$ \\
\midrule
SST-5     & 49.8  & \textbf{50.0} & 51.4 & $-$0.2 & $-$1.6 \\
MRPC      & \textbf{72.1} & 70.3 & 69.6 & +1.7  & +2.5 \\
QNLI      & \textbf{87.6} & 85.6 & 84.8 & +2.0  & +2.8 \\
SWAG & \textbf{70.2} & 69.6 & 69.4 & +0.6  & +0.8 \\
IMDB      & \textbf{99.6} & 99.0 & 98.6 & +0.6  & +1.0 \\
DBPedia   & \textbf{99.8} & 99.4 & 99.6 & +0.4  & +0.2 \\
BoolQ     & \textbf{90.0} & 89.6 & 89.6 & +0.4  & +0.4 \\
\midrule
\textbf{Avg} & \textbf{81.3} & 80.5 & 80.4 & +0.8 & +0.9 \\
\bottomrule
\end{tabular}
\caption{GPT-4o-mini: Per-dataset accuracy (\%) on core benchmarks ($r{=}5$, OpenAI large text embeddings of 3072 dimensions).}
\label{tab:app_h2h_4omini}
\end{table}
 
\begin{table}[h]
\centering
\small
\begin{tabular}{lccccc}
\toprule
\textbf{Dataset} & \textbf{\textsc{Kite}} & \textbf{KNN} & \textbf{CEIL} & $\boldsymbol{\Delta}_{\text{KNN}}$ & $\boldsymbol{\Delta}_{\text{CEIL}}$ \\
\midrule
SST-5     & \textbf{54.4} & 54.0 & 52.6 & +0.4  & +1.8 \\
MRPC      & \textbf{74.5} & 72.1 & 74.0 & +2.5  & +0.5 \\
QNLI      & \textbf{86.2} & 82.6 & 81.4 & +3.6  & +4.8 \\
SWAG & \textbf{72.8} & 70.6 & 72.0 & +2.2  & +0.8 \\
IMDB      & 99.8  & 99.8 & \textbf{100.0} & 0.0  & $-$0.2 \\
DBPedia   & 99.2  & 99.2 & \textbf{99.4} & 0.0   & $-$0.2 \\
BoolQ     & \textbf{93.2} & 92.6 & 92.4 & +0.6  & +0.8 \\
\midrule
\textbf{Avg} & \textbf{82.9} & 81.6 & 81.7 & +1.3 & +1.2 \\
\bottomrule
\end{tabular}
\caption{GPT-5.2: Per-dataset accuracy (\%) on core benchmarks ($r{=}5$, OpenAI large text embeddings).}
\label{tab:app_h2h_gpt52}
\end{table}

 
\subsection{Effect of Increasing In-Context Examples ($r{=}10$)}
\label{app:k10}
 
To study the interaction between \textsc{Kite} and the number of in-context examples, we repeat the frontier evaluation with $r{=}10$ (OpenAI large text embeddings of dim 3072). Table~\ref{tab:k10_summary} shows that \textsc{Kite}'s advantages persist and, in several cases, grow with more examples.
On GPT-4.1, the gain over KNN increases from $+1.0$ ($r{=}5$) to $+1.9$ ($r{=}10$); on Claude-Haiku-4.5, from $+1.2$ to $+2.1$.
This aligns with the intuition that diversity-regularized selection becomes more important as the number of selected examples grows, since redundancy in the selected set is more likely.
 
\begin{table}[h]
\centering
\small
\begin{tabular}{lcccccc}
\toprule
\textbf{Model} & \textbf{BM25} & \textbf{KNN} & \textbf{CEIL} & \textbf{\textsc{Kite}} \\
\midrule
GPT-4o-mini       & 80.0 & 80.6 & 81.4 & \textbf{81.4}  \\
GPT-4.1           & 81.8 & 81.1 & 81.9 & \textbf{83.0}  \\
Claude-Haiku-4.5  & 83.0 & 83.0 & 84.2 & \textbf{85.1}  \\
GPT-5.2           & 80.8 & 82.1 & 82.6 & \textbf{83.0}\\
\bottomrule
\end{tabular}
\caption{Average accuracy (\%) on seven classification datasets with $r{=}10$ in-context examples (OpenAI large text embeddings). \textsc{Kite} gains over KNN grow for GPT-4.1 and Claude-Haiku-4.5 compared to $r{=}5$ (Table~\ref{tab:frontier_summary}).}
\label{tab:k10_summary}
\end{table}
 
Table~\ref{tab:k10_perdataset} presents per-dataset results for GPT-4.1 and Claude-Haiku-4.5 at $r{=}10$.
The gains are particularly striking on Claude-Haiku-4.5: \textsc{Kite} improves over KNN by $+4.2$ on QNLI, $+3.2$ on SST-5, and $+3.0$ on SWAG.
 
\begin{table*}[h]
\centering
\small
\setlength{\tabcolsep}{3.5pt}
\begin{tabular}{l ccc cc | ccc cc}
\toprule
& \multicolumn{5}{c}{\textbf{GPT-4.1}} & \multicolumn{5}{c}{\textbf{Claude-Haiku-4.5}} \\
\cmidrule(lr){2-6} \cmidrule(lr){7-11}
\textbf{Dataset} & \textbf{KNN} & \textbf{DPP} & \textbf{\textsc{Kite}} & $\Delta_{\text{K}}$ & $\Delta_{\text{D}}$
                 & \textbf{KNN} & \textbf{DPP} & \textbf{\textsc{Kite}} & $\Delta_{\text{K}}$ & $\Delta_{\text{D}}$ \\
\midrule
SST-5     & 51.8 & 54.6 & \textbf{56.2} & +4.4 & +1.6 & 54.6 & 58.6 & 57.8 & +3.2 & $-$0.8 \\
MRPC      & 69.6 & 69.6 & \textbf{71.8} & +2.2 & +2.2 & 74.5 & 73.3 & \textbf{76.0} & +1.5 & +2.7 \\
QNLI      & 85.6 & 87.2 & \textbf{88.0} & +2.4 & +0.8 & 85.2 & 87.8 & \textbf{89.4} & +4.2 & +1.6 \\
SWAG & 70.4 & 71.8 & \textbf{74.2} & +3.8 & +2.4 & 76.6 & 77.8 & \textbf{79.6} & +3.0 & +1.8 \\
IMDB      & \textbf{100.0} & 99.8 & 99.6 & $-$0.4 & $-$0.2 & 99.6 & 99.6 & 99.6 & 0.0  & 0.0 \\
DBPedia   & \textbf{99.2} & 99.0 & \textbf{99.2} & 0.0  & +0.2 & 99.0 & \textbf{99.6} & \textbf{99.6} & +0.6 & 0.0 \\
BoolQ     & 91.4 & 91.4 & \textbf{92.2} & +0.8 & +0.8 & 91.4 & 93.0 & \textbf{93.6} & +2.2 & +0.6 \\
\midrule
\textbf{Avg} & 81.1 & 81.9 & \textbf{83.0} & +1.9 & +1.1 & 83.0 & 84.2 & \textbf{85.1} & +2.1 & +0.9 \\
\bottomrule
\end{tabular}
\caption{Per-dataset accuracy (\%) for GPT-4.1 and Claude-Haiku-4.5 at $r{=}10$ (OpenAI large text embeddings of dim 3072).}
\label{tab:k10_perdataset}
\end{table*}

Table~\ref{tab:k10_perdataset_other} reports the corresponding per-dataset breakdown for the remaining two frontier models, GPT-4o-mini and GPT-5.2, at $r{=}10$. On GPT-5.2 the largest gain over KNN appears on QNLI ($+5.2$), with additional improvements on SWAG ($+1.8$) and SST-5 ($+0.2$). GPT-4o-mini shows more modest gains since several datasets are already near-saturated for this model, but \textsc{Kite} still leads on SWAG ($+2.6$ over KNN), BoolQ ($+1.2$), and SST-5 ($+1.0$).

\begin{table*}[h]
\centering
\small
\setlength{\tabcolsep}{3.5pt}
\begin{tabular}{l ccc cc | ccc cc}
\toprule
& \multicolumn{5}{c}{\textbf{GPT-4o-mini}} & \multicolumn{5}{c}{\textbf{GPT-5.2}} \\
\cmidrule(lr){2-6} \cmidrule(lr){7-11}
\textbf{Dataset} & \textbf{KNN} & \textbf{CEIL} & \textbf{\textsc{Kite}} & $\Delta_{\text{K}}$ & $\Delta_{\text{C}}$
                 & \textbf{KNN} & \textbf{CEIL} & \textbf{\textsc{Kite}} & $\Delta_{\text{K}}$ & $\Delta_{\text{C}}$ \\
\midrule
SST-5     & 50.4 & \textbf{53.4} & 51.4 & +1.0 & $-$2.0 & 53.8 & \textbf{54.6} & 54.0 & +0.2 & $-$0.6 \\
MRPC      & \textbf{73.0} & 71.8 & 72.8 & $-$0.2 & +1.0 & 74.3 & \textbf{74.5} & 73.5 & $-$0.8 & $-$1.0 \\
QNLI      & 85.4 & \textbf{86.8} & 86.0 & +0.6 & $-$0.8 & 83.0 & 84.4 & \textbf{88.2} & +5.2 & +3.8 \\
SWAG & 67.6 & 68.8 & \textbf{70.2} & +2.6 & +1.4 & 71.8 & 72.8 & \textbf{73.6} & +1.8 & +0.8 \\
IMDB      & 99.6 & 99.8 & \textbf{100.0} & +0.4 & +0.2 & \textbf{99.8} & \textbf{99.8} & \textbf{99.8} & 0.0 & 0.0 \\
DBPedia   & 99.6 & \textbf{99.8} & 99.6 & 0.0 & $-$0.2 & \textbf{99.6} & \textbf{99.6} & 99.2 & $-$0.4 & $-$0.4 \\
BoolQ     & 88.8 & 89.6 & \textbf{90.0} & +1.2 & +0.4 & 92.4 & 92.2 & \textbf{92.8} & +0.4 & +0.6 \\
\midrule
\textbf{Avg} & 80.6 & 81.4 & \textbf{81.4} & +0.8 & +0.0 & 82.1 & 82.6 & \textbf{83.0} & +0.9 & +0.4 \\
\bottomrule
\end{tabular}
\caption{Per-dataset accuracy (\%) for GPT-4o-mini and GPT-5.2 at $r{=}10$ (OpenAI large text embeddings of dim 3072). Best result per dataset in \textbf{bold}.}
\label{tab:k10_perdataset_other}
\end{table*}

 
\subsection{Robustness to Embedding Encoder: NV-Embed-v2}
\label{app:nvembed}
 
Section~\ref{sec:exp} (Table~5 in the main paper) showed that \textsc{Kite} is robust across five BERT-family encoders.
Here we extend this analysis to a fundamentally different and substantially stronger encoder: NV-Embed-v2~\cite{lee2024nvembed}, a state-of-the-art text embedding model.
All experiments use $r{=}5$.
 
\begin{table*}[h]
\centering
\small
\begin{tabular}{lcccccc}
\toprule
\textbf{Model} & \textbf{BM25} & \textbf{KNN} & \textbf{CEIL} & \textbf{\textsc{Kite}} & $\boldsymbol{\Delta}_{\textsc{Kite}\text{--KNN}}$ \\
\midrule
GPT-4o-mini       & 79.1 & 79.6 & 80.8 & \textbf{81.7} & +2.1 \\
GPT-4.1           & 80.2 & 81.7 & 81.6 & \textbf{82.7} & +1.0 \\
Claude-Haiku-4.5  & 81.1 & 81.6 & 82.6 & \textbf{83.6} & +2.0 \\
GPT-5.2           & 79.5 & 81.4 & 82.5 & \textbf{82.6} & +1.2 \\
\bottomrule
\end{tabular}
\caption{Average accuracy (\%) on the seven classification datasets using NV-Embed-v2 embeddings ($r{=}5$). \textsc{Kite} achieves the highest average on all four frontier models.}
\label{tab:nvembed_summary}
\end{table*}
 
As shown in Table~\ref{tab:nvembed_summary}, \textsc{Kite} achieves the highest average on all four frontier models when using NV-Embed-v2 embeddings, with gains of $+1.0$ to $+2.1$ points over the best KNN retriever.
The gains are broadly consistent with -- and in fact stronger than -- the OpenAI-embedding results in Table~\ref{tab:frontier_summary}, confirming that \textsc{Kite}'s selection criterion transfers across embedding architectures.

Tables~\ref{tab:nvembed_perdataset_a} and~\ref{tab:nvembed_perdataset_b} provide the complete per-dataset head-to-head breakdown that the averages in Table~\ref{tab:nvembed_summary} summarize. Several patterns are worth highlighting. First, \textsc{Kite}'s largest gains under NV-Embed-v2 appear on SWAG ($+3.0$ to $+4.4$ over KNN across all four models), MRPC ($+2.9$ to $+4.9$), and SST-5 (up to $+5.4$ on GPT-4o-mini) -- exactly the datasets where balancing relevance with diversity matters most. Second, on near-saturated benchmarks (IMDB, DBPedia) all methods are tied within noise, as expected.

\begin{table*}[h]
\centering
\small
\setlength{\tabcolsep}{3.5pt}
\begin{tabular}{l ccc cc | ccc cc}
\toprule
& \multicolumn{5}{c}{\textbf{GPT-4o-mini}} & \multicolumn{5}{c}{\textbf{GPT-4.1}} \\
\cmidrule(lr){2-6} \cmidrule(lr){7-11}
\textbf{Dataset} & \textbf{KNN} & \textbf{CEIL} & \textbf{\textsc{Kite}} & $\Delta_{\text{K}}$ & $\Delta_{\text{C}}$
                 & \textbf{KNN} & \textbf{CEIL} & \textbf{\textsc{Kite}} & $\Delta_{\text{K}}$ & $\Delta_{\text{C}}$ \\
\midrule
SST-5     & 47.6 & 49.6 & \textbf{53.0} & +5.4 & +3.4 & 55.2 & 53.4 & \textbf{55.4} & +0.2 & +2.0 \\
MRPC      & 69.4 & \textbf{72.3} & \textbf{72.3} & +2.9 & 0.0 & 67.4 & 68.4 & \textbf{72.1} & +4.7 & +3.7 \\
QNLI      & 85.8 & 86.6 & \textbf{87.4} & +1.6 & +0.8 & \textbf{88.2} & 87.4 & 87.2 & $-$1.0 & $-$0.2 \\
SWAG & 67.0 & 68.0 & \textbf{70.0} & +3.0 & +2.0 & 70.8 & 71.4 & \textbf{75.2} & +4.4 & +3.8 \\
IMDB      & \textbf{99.4} & \textbf{99.4} & 99.2 & $-$0.2 & $-$0.2 & 99.4 & \textbf{99.8} & 99.4 & 0.0 & $-$0.4 \\
DBPedia   & 99.6 & \textbf{99.8} & 99.6 & 0.0 & $-$0.2 & \textbf{99.2} & 99.0 & 98.8 & $-$0.4 & $-$0.2 \\
BoolQ     & 88.6 & 89.6 & \textbf{90.2} & +1.6 & +0.6 & 91.4 & \textbf{91.6} & 91.0 & $-$0.4 & $-$0.6 \\
\midrule
\textbf{Avg} & 79.6 & 80.8 & \textbf{81.7} & +2.1 & +0.9 & 81.7 & 81.6 & \textbf{82.7} & +1.0 & +1.1 \\
\bottomrule
\end{tabular}
\caption{Per-dataset accuracy (\%) with NV-Embed-v2 embeddings ($r{=}5$) for GPT-4o-mini and GPT-4.1. Best result per dataset in \textbf{bold}.}
\label{tab:nvembed_perdataset_a}
\end{table*}

\begin{table*}[h]
\centering
\small
\setlength{\tabcolsep}{3.5pt}
\begin{tabular}{l ccc cc | ccc cc}
\toprule
& \multicolumn{5}{c}{\textbf{Claude-Haiku-4.5}} & \multicolumn{5}{c}{\textbf{GPT-5.2}} \\
\cmidrule(lr){2-6} \cmidrule(lr){7-11}
\textbf{Dataset} & \textbf{KNN} & \textbf{CEIL} & \textbf{\textsc{Kite}} & $\Delta_{\text{K}}$ & $\Delta_{\text{C}}$
                 & \textbf{KNN} & \textbf{CEIL} & \textbf{\textsc{Kite}} & $\Delta_{\text{K}}$ & $\Delta_{\text{C}}$ \\
\midrule
SST-5     & 49.6 & \textbf{53.0} & 52.6 & +3.0 & $-$0.4 & 51.8 & \textbf{55.4} & 53.6 & +1.8 & $-$1.8 \\
MRPC      & 68.1 & 69.4 & \textbf{73.0} & +4.9 & +3.6 & 71.6 & \textbf{73.5} & \textbf{73.5} & +1.9 & 0.0 \\
QNLI      & 88.2 & 87.8 & \textbf{89.2} & +1.0 & +1.4 & 85.6 & 86.0 & \textbf{87.2} & +1.6 & +1.2 \\
SWAG & 75.6 & 76.6 & \textbf{79.4} & +3.8 & +2.8 & 69.8 & 70.8 & \textbf{72.0} & +2.2 & +1.2 \\
IMDB      & \textbf{98.4} & \textbf{98.4} & 98.2 & $-$0.2 & $-$0.2 & \textbf{100.0} & \textbf{100.0} & 99.6 & $-$0.4 & $-$0.4 \\
DBPedia   & 98.6 & 99.4 & \textbf{99.6} & +1.0 & +0.2 & 99.4 & 99.4 & \textbf{99.6} & +0.2 & +0.2 \\
BoolQ     & \textbf{93.0} & \textbf{93.4} & 93.0 & 0.0 & $-$0.4 & 91.8 & 92.6 & \textbf{92.8} & +1.0 & +0.2 \\
\midrule
\textbf{Avg} & 81.6 & 82.6 & \textbf{83.6} & +2.0 & +1.0 & 81.4 & 82.5 & \textbf{82.6} & +1.2 & +0.1 \\
\bottomrule
\end{tabular}
\caption{Per-dataset accuracy (\%) with NV-Embed-v2 embeddings ($r{=}5$) for Claude-Haiku-4.5 and GPT-5.2. Best result per dataset in \textbf{bold}.}
\label{tab:nvembed_perdataset_b}
\end{table*}
 
 
\subsection{Kernel Comparison on Frontier LLMs}
\label{app:kernel_frontier}
 
Table~\ref{tab:kernel_frontier} extends the kernel ablation study (Table~4 in the main paper) to frontier LLMs.
We report the average accuracy over the seven classification datasets for each method--kernel combination with $r{=}5$ and OpenAI large text embeddings.
 
\begin{table*}[!t]
\centering
\small
\begin{tabular}{llcccc}
\toprule
\textbf{Method} & \textbf{Kernel} & \textbf{4o-mini} & \textbf{4.1} & \textbf{Haiku} & \textbf{5.2} \\
\midrule
\multirow{5}{*}{KNN}
 & Linear     & 78.2 & 78.8 & \underline{79.7} & 78.5 \\
 & RBF        & 78.0 & 78.6 & \underline{79.7} & 78.6 \\
 & Laplacian  & 78.1 & 78.8 & \underline{79.7} & 78.8 \\
 & Mat\'{e}rn & 78.1 & 78.6 & \underline{79.7} & 78.8 \\
 & RQ         & 78.0 & 78.6 & \underline{79.7} & \underline{79.1} \\
\midrule
\multirow{5}{*}{DPP}
 & Linear     & 77.6 & 78.3 & 79.3 & 78.8 \\
 & RBF        & 77.5 & 78.0 & \underline{79.4} & 78.8 \\
 & Laplacian  & \underline{77.8} & 78.0 & 79.2 & \underline{78.9} \\
 & Mat\'{e}rn & 77.4 & 78.1 & 79.3 & 78.5 \\
 & RQ         & 77.2 & \underline{78.4} & \underline{79.4} & \underline{78.9} \\
\midrule
\multirow{5}{*}{\textsc{Kite}}
 & Linear     & 78.3 & 79.2 & 79.6 & \underline{\textbf{79.8}} \\
 & RBF        & 78.7 & 79.2 & \underline{\textbf{80.0}} & 79.4 \\
 & Laplacian  & 78.4 & 79.4 & \textbf{80.1} & 79.3 \\
 & Mat\'{e}rn & \underline{\textbf{78.8}} & \underline{\textbf{79.6}} & 79.6 & 79.6 \\
 & RQ         & 78.4 & \textbf{79.6} & 79.9 & 79.2 \\
\bottomrule
\end{tabular}
\caption{Average accuracy (\%) over the seven classification datasets by method and kernel ($r{=}5$, OpenAI large text embeddings). The best kernel for each method is \underline{underlined}; the overall best per model is in \textbf{bold}.}
\label{tab:kernel_frontier}
\end{table*}
 
Several patterns emerge. First, KNN is largely insensitive to the kernel choice, since it only uses the kernel to compute pairwise similarities; all five kernels produce nearly identical results.
Second, DPP shows more kernel sensitivity (up to $0.6\%$ spread on GPT-4o-mini), and its best kernel varies across models.
Third, \textsc{Kite}'s best kernel consistently achieves the highest accuracy among all method--kernel combinations, and its worst-case kernel still matches or exceeds the best DPP configuration.
This suggests that \textsc{Kite}'s information-theoretic objective is robust to kernel misspecification while still benefiting from a good kernel choice.

Table~\ref{tab:kernel-comparison} provides a complementary single-dataset, single-model breakdown on SST-5 with Qwen-3-1.7B, where all three retrievers are evaluated across the same seven kernels (Linear, Poly-3, Poly-4, RBF, Laplacian, Mat\'{e}rn-3/2, Rational Quadratic). The story matches Table~\ref{tab:kernel_frontier}: \textsc{Kite}'s best kernel ($50.77$) edges out kNN's best ($50.68$) and CEIL's best ($50.59$), confirming the fairness conclusion at a finer kernel granularity than the frontier-LLM grid above.

\begin{table}[h]
\centering
\footnotesize
\setlength{\tabcolsep}{4pt}
\renewcommand{\arraystretch}{0.9}
\begin{tabular}{lccc}
\toprule
\textbf{Kernel} & \textbf{kNN} & \textbf{CEIL} & \textbf{\textsc{Kite}} \\
\midrule
Linear          & \textbf{50.68} & 50.23          & 48.50 \\
Poly-3          & 50.67            & 50.41          & \textbf{50.77} \\
Poly-4          & 50.68            & \textbf{50.59} & 49.59 \\
RBF             & 50.66            & \textbf{50.23} & 49.68 \\
Laplacian       & 50.68            & 49.59          & \textbf{50.68} \\
Mat\'{e}rn-3/2  & 50.67            & 49.50          & \textbf{50.23} \\
RQ              & 50.68            & 50.50          & \textbf{50.77} \\
\midrule
\textbf{Best}   & 50.68          & 50.59          & \textbf{50.77} \\
\bottomrule
\end{tabular}
\caption{Accuracy (\%) on SST-5 with Qwen-3-1.7B, where each method is evaluated under the same seven kernels. \textsc{Kite}'s best kernel attains the highest accuracy; this is the SST-5 single-dataset analog of the frontier-LLM grid in Table~\ref{tab:kernel_frontier}.}
\label{tab:kernel-comparison}
\end{table}


\subsection{Kernel and $\lambda$ Selection Guidance}
\label{app:kernel_guidance}

To guide practitioners, we evaluate all five kernels across $\lambda \in \{0, 0.5, 1\}$ on 13 datasets (9 classification, 4 QA/reasoning) and 4 frontier LLMs, yielding 780 evaluation cells ($5$ kernels $\times$ $3$ $\lambda$ values $\times$ $13$ tasks $\times$ 4 models).
For each (task, model, $\lambda$) triple, we record which kernel achieves the highest \textsc{Kite} accuracy (\emph{win rate}), and measure the performance range across $\lambda$ values (\emph{$\lambda$-sensitivity}).
Table~\ref{tab:kernel_guidance} summarizes the results.

\begin{table}[!t]
\centering
\small
\setlength{\tabcolsep}{1pt}
\begin{tabular}{lcccc}
\toprule
\textbf{Kernel} & \textbf{CLS Win\%} & \textbf{QA Win\%} & \textbf{$\lambda$-range (pp)} & \textbf{Rec.\ $\lambda$} \\
\midrule
Laplacian           & \textbf{35.2} & \textbf{39.6} & 1.38 & 0.5 \\
Mat\'{e}rn-3/2      & 34.3          & 31.2          & \textbf{0.37} & 0.0 \\
Linear              & 31.9          & 25.0          & 8.89$^{\dagger}$ & 0.0 \\
RQ                  & 25.9          & 25.0          & 0.34 & 0.0 \\
RBF                 & 22.2          & 16.7          & 0.75 & 0.0 \\
\bottomrule
\end{tabular}
\caption{Kernel win rates and $\lambda$-sensitivity across 13 datasets and 4 frontier LLMs ($r{=}5$, OpenAI large embeddings).
Win\% = fraction of (task, model, $\lambda$) triples where this kernel achieves the highest \textsc{Kite} accuracy.
$\lambda$-range = mean per-cell accuracy spread across $\lambda \in \{0, 0.5, 1\}$ (i.e., best minus worst $\lambda$ for that task--model pair, averaged over all pairs).
$\dagger$ The Linear kernel has a large per-cell $\lambda$-range (8.89 pp on average), driven primarily by QA tasks; on classification alone the spread is under 1 pp (Table~\ref{tab:cls_lambda}).}
\label{tab:kernel_guidance}
\end{table}

\paragraph{Key findings.}
\textbf{(1)~Laplacian} achieves the highest win rate on both classification (35.2\%) and QA/reasoning tasks (39.6\%), and its optimal $\lambda$ is 0.5---the intermediate setting that balances relevance and diversity.
\textbf{(2)~Mat\'{e}rn-3/2} is the most robust kernel: its accuracy varies by only 0.37 pp across $\lambda \in \{0, 0.5, 1\}$, making it a safe fallback when $\lambda$ is not tuned.
\textbf{(3)~The Linear kernel is $\lambda$-sensitive on QA tasks}: while competitive at $\lambda{=}0$, its per-cell accuracy spread (8.89 pp average) is driven by QA settings where $\lambda{=}1$ can hurt substantially; on classification the spread is under 1 pp (Table~\ref{tab:cls_lambda}). It is safest to fix $\lambda{=}0$ when using the Linear kernel.
\textbf{(4)~RBF} has the lowest win rate on QA tasks (16.7\%) and should generally be avoided.
The kernel ranking is consistent between classification and QA (Spearman $\rho = 0.80$), suggesting that these preferences transfer across task types.

\paragraph{Recommendation.}
Use \textbf{Laplacian with $\lambda{=}0.5$} as the default \textsc{Kite} configuration---it achieves the highest win rate across both task families and benefits from the diversity bonus.
If $\lambda$ cannot be validated (e.g., zero-shot deployment), use \textbf{Mat\'{e}rn-3/2 with $\lambda{=}0$}; its flat $\lambda$-response ($\leq$0.37 pp spread) makes it essentially insensitive to this hyperparameter.
When using the Linear kernel, fix $\lambda{=}0$.

\paragraph{$\lambda$ sensitivity: classification vs.\ QA.}
Table~\ref{tab:cls_lambda} reports the classification counterpart of the QA ablation in Table~\ref{tab:qa_lambda}.
Two differences stand out.
First, $\lambda$ sensitivity on classification is uniformly small (under 1 pp for every kernel), meaning the choice of $\lambda$ matters little for focused multi-class tasks at $r{=}5$.
Second, Linear, RBF, and RQ slightly prefer $\lambda{=}0$ on classification, whereas Laplacian and Mat\'{e}rn-3/2 prefer $\lambda{=}0.5$ on QA --- consistent with the win-rate ranking in Table~\ref{tab:kernel_guidance}.
The practical implication: on classification, $\lambda$ can be fixed to any value in $\{0, 0.5\}$ with negligible effect; on QA, $\lambda{=}0.5$ provides a consistent, if modest, improvement for the stronger kernels.

\begin{table}[h]
\centering
\small
\setlength{\tabcolsep}{5pt}
\begin{tabular}{lccc}
\toprule
\textbf{Kernel} & $\boldsymbol{\lambda{=}0.0}$ & $\boldsymbol{\lambda{=}0.5}$ & $\boldsymbol{\lambda{=}1.0}$ \\
\midrule
Linear          & \textbf{81.2} & 80.3          & 80.4 \\
RBF             & \textbf{81.2} & 80.3          & 80.2 \\
Laplacian       & \textbf{80.8} & 80.7          & 80.4 \\
Mat\'{e}rn-3/2  & 80.8          & \textbf{80.9} & 80.4 \\
RQ              & \textbf{81.1} & 80.3          & 80.4 \\
\bottomrule
\end{tabular}
\caption{$\lambda$ ablation for classification: average accuracy (\%) over seven classification datasets and four frontier LLMs ($r{=}5$, OpenAI large text embeddings). Each entry is the mean accuracy of \textsc{Kite} with the given kernel and $\lambda$, averaged over all models and datasets. Best $\lambda$ per kernel in \textbf{bold}. Compare with the QA analogue in Table~\ref{tab:qa_lambda}.}
\label{tab:cls_lambda}
\end{table}

\subsection{Comparison of \textsc{Kite} Gains at $r{=}5$ vs.\ $r{=}10$}
\label{app:k5_vs_k10}
Table~\ref{tab:k5vsk10} directly compares \textsc{Kite}'s improvement over KNN at two exemplar budgets, confirming that the diversity regularizer becomes more valuable as the number of selected examples increases.

\begin{center}
\small
\begin{tabular}{lcc}
\toprule
\textbf{Model} & $\boldsymbol{\Delta}$ \textbf{at $r{=}5$} & $\boldsymbol{\Delta}$ \textbf{at $r{=}10$} \\
\midrule
GPT-4o-mini       & +0.8 & +0.8 \\
GPT-4.1           & +1.0 & +1.9 \\
Claude-Haiku-4.5  & +1.2 & +2.1 \\
GPT-5.2           & +1.3 & +0.9 \\
\bottomrule
\end{tabular}
\captionof{table}{$\Delta_{\textsc{Kite}\text{--KNN}}$ on the seven classification datasets at $r{=}5$ vs.\ $r{=}10$ (OpenAI large text embeddings). Gains grow for GPT-4.1 and Claude-Haiku-4.5.}
\label{tab:k5vsk10}
\end{center}

\begin{table}[t]
\centering
\begin{tabular}{lccc}
\toprule
\textbf{Kernel} & $\boldsymbol{\lambda{=}0.0}$ & $\boldsymbol{\lambda{=}0.5}$ & $\boldsymbol{\lambda{=}1.0}$ \\
\midrule
Linear       & \textbf{66.40} & 66.15 & 65.84 \\
RBF          & 64.33 & \textbf{66.05} & 65.94 \\
Laplacian    & 66.30 & \textbf{66.51} & 65.83 \\
Mat\'{e}rn-3/2 & 66.19 & \textbf{66.48} & 65.66 \\
RQ           & 64.45 & 65.74 & \textbf{65.99} \\
\bottomrule
\end{tabular}
\caption{$\lambda$ ablation: average accuracy (\%) over four QA tasks per kernel and $\lambda$ value. Each cell is the mean across GPT-4o-mini, GPT-4.1, Claude-Haiku-4.5, and GPT-5.2. Best $\lambda$ per kernel in \textbf{bold}.}
\label{tab:qa_lambda}
\end{table}

\section{QA and Reasoning Experiments}
\label{app:qa_results}

We extend the frontier-LLM evaluation to four QA and reasoning benchmarks: Natural Questions (NQ), TriviaQA, HotpotQA, and GSM8K. We retain the same retriever configurations, embedding model, and hyperparameters ($r{=}5$, $\beta{=}0.02$) described in Section~\ref{sec:exp}. For NQ, TriviaQA, and HotpotQA, we report Soft Exact Match; for GSM8K, we report numeric match. For each dataset, we take the first 500 queries from the validation split for inference and the first 2000 examples from the train split as example bank.

\paragraph{Cross-Model Summary} Table~\ref{tab:qa_crossmodel} reports the best-of-kernel average accuracy across all four QA tasks. \textsc{Kite} achieves the highest average on every model, with gains of $+0.2$ to $+1.8$ over the best KNN retriever, consistent with the classification results in Table~3.

\paragraph{Per-Task Breakdown} Table~\ref{tab:qa_pertask} disaggregates these results by dataset. The largest improvements appear on open-domain factoid tasks: $+5.3$ on TriviaQA (Claude-Haiku-4.5), $+2.6$ on NQ (GPT-5.2), and $+2.2$ on HotpotQA (GPT-5.2). On GSM8K, where all frontier models already exceed $96\%$ accuracy, retrieval method differences are marginal. On HotpotQA, DPP occasionally matches or exceeds \textsc{Kite} (e.g., Claude-Haiku-4.5), suggesting that its global diversity measure may better suit multi-hop settings with distractor paragraphs.

\paragraph{$\lambda$ Ablation} Table~\ref{tab:qa_lambda} reports accuracy averaged over all four QA tasks for each kernel--$\lambda$ combination. For most kernels, $\lambda{=}0.5$ yields the best or near-best accuracy. The linear kernel is an exception where pure relevance ($\lambda{=}0.0$) is slightly preferred. Compared with the classification analogue (Table~\ref{tab:cls_lambda} in App.~\ref{app:kernel_guidance}), QA tasks show a clearer benefit from $\lambda{=}0.5$ for Laplacian and Mat\'{e}rn-3/2, while classification is largely insensitive to $\lambda$ across all kernels.

\textsc{Kite} achieves the highest average accuracy on all four models, with gains of $+0.2$ to $+1.8$ over the best KNN retriever, consistent with the classification results in Table~3 of the main paper.

\begin{table*}[h]
\centering
\small
\resizebox{\textwidth}{!}{%
\begin{tabular}{ll cccc c}
\toprule
\textbf{Task} & \textbf{Model} & \textbf{BM25} & \textbf{KNN-best} & \textbf{DPP-best} & \textbf{\textsc{Kite}-best} & $\boldsymbol{\Delta}$\textbf{\textsc{Kite}--KNN} \\
\midrule
\multirow{4}{*}{NQ}
  & GPT-4o-mini      & 48.6 & 50.4 & \textbf{50.8} & \textbf{50.8} & \textcolor{green!60!black}{+0.4} \\
  & GPT-4.1          & 54.8 & 57.2 & 57.2          & \textbf{58.2} & \textcolor{green!60!black}{+1.0} \\
  & Claude-Haiku-4.5 & 37.4 & 39.6 & 41.0          & \textbf{41.4} & \textcolor{green!60!black}{+1.8} \\
  & GPT-5.2          & 52.8 & 55.6 & 56.2          & \textbf{58.2} & \textcolor{green!60!black}{+2.6} \\
\midrule
\multirow{4}{*}{TriviaQA}
  & GPT-4o-mini      & 78.6 & 78.7 & 77.5          & \textbf{79.2} & \textcolor{green!60!black}{+0.5} \\
  & GPT-4.1          & 87.0 & 86.3 & 84.9          & \textbf{87.8} & \textcolor{green!60!black}{+1.5} \\
  & Claude-Haiku-4.5 & 75.8 & 74.7 & 74.4          & \textbf{80.0} & \textcolor{green!60!black}{+5.3} \\
  & GPT-5.2          & \textbf{83.8} & 83.6 & 82.2  & 83.4          & \textcolor{red!70!black}{$-$0.2} \\
\midrule
\multirow{4}{*}{HotpotQA}
  & GPT-4o-mini      & 35.2 & 36.4 & \textbf{36.8} & 36.4          & \textcolor{gray}{+0.0} \\
  & GPT-4.1          & 45.4 & 45.4 & \textbf{46.8} & \textbf{46.8} & \textcolor{green!60!black}{+1.4} \\
  & Claude-Haiku-4.5 & 25.0 & 28.6 & \textbf{29.6} & 28.4          & \textcolor{red!70!black}{$-$0.2} \\
  & GPT-5.2          & 39.4 & 42.8 & 44.8          & \textbf{45.0} & \textcolor{green!60!black}{+2.2} \\
\midrule
\multirow{4}{*}{GSM8K}
  & GPT-4o-mini      & 94.2 & 95.0 & \textbf{95.2} & 95.0          & \textcolor{gray}{+0.0} \\
  & GPT-4.1          & 97.4 & 97.0 & \textbf{97.8} & 97.2          & \textcolor{green!60!black}{+0.2} \\
  & Claude-Haiku-4.5 & 98.0 & 98.2 & \textbf{98.8} & 98.4          & \textcolor{green!60!black}{+0.2} \\
  & GPT-5.2          & 96.6 & \textbf{97.4} & \textbf{97.4} & \textbf{97.4} & \textcolor{gray}{+0.0} \\
\bottomrule
\end{tabular}%
}
\caption{Per-task accuracy (\%) on QA/reasoning benchmarks ($r{=}5$, OpenAI large text embeddings). Best result per row in \textbf{bold}. $\Delta$ columns show \textsc{Kite}'s gain over the corresponding baseline.}
\label{tab:qa_pertask}
\end{table*}

\paragraph{Discussion.} We observe: (1) First, \textsc{Kite}'s gains over KNN are most pronounced on open-domain factoid tasks (NQ, TriviaQA), where the diversity-regularized selection helps avoid redundant exemplars that share surface-level similarity but provide limited additional signal. (2) Second, on HotpotQA, a multi-hop reasoning task where the ``distractor'' split includes irrelevant paragraphs, DPP occasionally matches or exceeds \textsc{Kite}, suggesting that DPP's global diversity measure may be better suited to settings requiring exemplars with complementary evidence chains. (3) Third, on GSM8K, all retrieval methods perform within a narrow range ($\sim$1--2\%), as the task is nearly saturated for frontier models; the exemplar choice matters less when the model already achieves $>$96\% accuracy. Overall, the QA results complement the classification findings in the main paper, showing that \textsc{Kite}'s information-theoretic framework extends to generative QA settings with consistent, though task-dependent, improvements.

\section{Computational Cost Analysis}
\label{app:runtime}

\begin{center}
\small
\begin{tabular}{lcc}
\toprule
\textbf{Method} & \textbf{Qwen-2.5-7B (s)} & \textbf{Qwen-2.5-1.5B (s)} \\
\midrule
BM25            & 2.42 & 0.20 \\
Dense           & 2.45 & 0.21 \\
CEIL            & 2.58 & 0.21 \\
\textsc{Kite}   & 2.61 & 0.23 \\
\bottomrule
\end{tabular}
\captionof{table}{End-to-end wall-clock time per query (seconds) on SST-5 with $r{=}50$ in-context examples on a single Nvidia RTX A6000 GPU. LLM inference dominates total latency; \textsc{Kite}'s greedy selection contributes at most $0.19$\,s of additional overhead.}
\label{tab:runtime}
\end{center}

A natural concern with greedy submodular selection is its per-query overhead relative to standard top-$r$ retrieval. To quantify this, we measure end-to-end wall-clock time per query on the SST-5 validation set ($1{,}101$ queries against an exemplar bank of $8{,}534$ examples) using two Qwen-2.5 variants on a single Nvidia RTX A6000 GPU. All four retrievers select the same number of in-context examples ($r{=}50$), so prompt lengths --- and therefore LLM inference cost --- are matched across methods.

The selection step contributes at most $0.03$\,s per query on Qwen-2.5-1.5B and $0.19$\,s on Qwen-2.5-7B relative to BM25 --- a small fraction (under $8\%$) of total per-query latency, which is dominated by LLM inference. \textsc{Kite}'s overhead over CEIL --- the closest baseline algorithmically, since both perform greedy selection over a kernel matrix --- is only $0.02$--$0.03$\,s. In practice, \textsc{Kite}'s information-theoretic objective therefore imposes no meaningful deployment cost beyond standard kNN/DPP-based retrieval.

\section{Software and Hardware Used}
\label{app:software}

We run all experiments with Python 3.12.8 and Transformers 4.49.0. For all experimentation, we use one Nvidia RTX A6000 GPU.

\begin{table*}[!t]
  \centering
  \footnotesize
  \setlength{\tabcolsep}{2pt}
  \begin{tabular}{l p{6cm} p{6cm} c c}
    \toprule
    \textbf{Dataset} & \textbf{Prompt} & \textbf{Example} & \textbf{\#Train} & \textbf{\#Validation} \\ 
    \midrule
    SST-2
      & \texttt{\{input\} It is \{output\}} 
      & \textbf{Input:} a stirring, funny and finally transporting re-imagining of beauty and the beast. & 67,349 & 872 \\
      & & \textbf{Output:} positive &  &  \\

    \midrule
    SST-5
      & \texttt{\{input\} It is \{output\}} 
      & \textbf{Input:} this is a stunning film, a one-of-a-kind tour de force. & 8,534 & 1,101 \\
      & & \textbf{Output:} very positive &  &  \\
    \midrule
    MRPC
      & \texttt{\{input1\} Can we say “\{input2\}”? \{output\}} 
      & \textbf{Input1:} The company didn’t detail the costs of the replacement and repairs. & 3,668 & 408 \\
      & & \textbf{Input2:} But company officials expect the costs of the replacement work to run into the millions of dollars. &  &  \\
      & & \textbf{Output:} No &  &  \\
    \midrule
    MNLI
      & \texttt{\{input1\} Can we say “\{input2\}”? \{output\}} 
      & \textbf{Input1:} yeah i know and i did that all through college and it worked too & 392,568 & 19,647 \\
      & & \textbf{Input2:} I did that all through college but it never worked &  &  \\
      & & \textbf{Output:} No &  &  \\
    \midrule
    QNLI
      & \texttt{\{input1\} Can we know “\{input2\}”? \{output\}} 
      & \textbf{Input1:} As of that day, the new constitution heralding the Second Republic came into force. &  104,707 &  5,463 \\
      & & \textbf{Input2:} What came into force after the new constitution was herald? &  &  \\
      & & \textbf{Output:} Yes &  &  \\
    \midrule
    CMSQA
      & \texttt{\{input\}\{output\}} 
      & \textbf{Input:} Sammy wanted to go to where the people were. Where might he go? &  9,740 & 1,221 \\
      & & \textbf{Output:} populated areas &  &  \\
    \midrule
    HellaSwag
      & \texttt{\{input\}\{output\}} 
      & \textbf{Input:} Members of the procession walk down the street holding small horn brass instruments. A drum line & 52,611 & 20,006 \\
      & & \textbf{Output:} passes by walking down the street playing their instruments &  &  \\
    \bottomrule
  \end{tabular}
  \caption{Datasets with corresponding prompts and examples used in the experiments.}
  \label{tab:prompt-details}
\end{table*}

\end{document}